\documentclass[runningheads]{llncs}
\usepackage{graphicx}

\usepackage{tikz}
\usetikzlibrary{arrows, backgrounds, fit, positioning, calc, shapes.symbols, shapes.callouts, patterns}
\tikzset{fit margins/.style={/tikz/afit/.cd,#1,
    /tikz/.cd,
    inner xsep=\pgfkeysvalueof{/tikz/afit/left}+\pgfkeysvalueof{/tikz/afit/right},
    inner ysep=\pgfkeysvalueof{/tikz/afit/top}+\pgfkeysvalueof{/tikz/afit/bottom},
    xshift=-\pgfkeysvalueof{/tikz/afit/left}+\pgfkeysvalueof{/tikz/afit/right},
    yshift=-\pgfkeysvalueof{/tikz/afit/bottom}+\pgfkeysvalueof{/tikz/afit/top}},
    afit/.cd,left/.initial=2pt,right/.initial=2pt,bottom/.initial=2pt,top/.initial=2pt}
\usepackage{comment}
\usepackage{amsmath,amssymb} 
\usepackage{color}
\usepackage[pagebackref,breaklinks,colorlinks]{hyperref}

\usepackage[accsupp]{axessibility}  

\usepackage{booktabs}
\usepackage{pifont}
\newcommand{\cmark}{\ding{51}}%
\newcommand{\xmark}{\ding{55}}%
\usepackage{multirow}
\usepackage{float}
\usepackage{wrapfig}

\usepackage[capitalize]{cleveref}
\crefname{section}{Sec.}{Secs.}
\Crefname{section}{Section}{Sections}
\Crefname{table}{Table}{Tables}
\crefname{table}{Tab.}{Tabs.}

\makeatletter
\renewcommand*{\@fnsymbol}[1]{\ensuremath{\ifcase#1\or \star\or \dagger\or \ddagger\or
   \mathsection\or \mathparagraph\or \|\or \star\star\or \dagger\dagger
   \or \ddagger\ddagger \else\@ctrerr\fi}}
\makeatother

\begin{document}
\pagestyle{headings}
\mainmatter
\def\ECCVSubNumber{1247}  

\title{CODA: A Real-World Road Corner Case Dataset for Object Detection in Autonomous Driving} 

\titlerunning{CODA: A Real-World Road Corner Case Dataset for Autonomous Driving}
%
\author{Kaican Li\inst{1}\thanks{Equal contribution.} \and
Kai Chen\inst{3}$^\star$ \and
Haoyu Wang\inst{1}$^\star$ \and
Lanqing Hong\inst{1}\thanks{Corresponding author at \href{mailto:honglanqing@huawei.com}{\texttt{honglanqing@huawei.com}}.} \and
Chaoqiang Ye\inst{1} \and
Jianhua Han\inst{1} \and
Yukuai Chen\inst{2} \and
Wei Zhang\inst{1} \and
Chunjing Xu\inst{1} \and
Dit-Yan Yeung\inst{3} \and
Xiaodan Liang\inst{5} \and
Zhenguo Li\inst{1} \and
Hang Xu\inst{1}}
\authorrunning{K. Li et al.}
%
\institute{Huawei Noah's Ark Lab \and
Huawei Intelligent Automotive Solution BU \and
Hong Kong University of Science and Technology \and
Sun Yat-sen University}


\maketitle

\begin{abstract}
    Contemporary deep-learning object detection methods for autonomous driving usually presume fixed categories of common traffic participants, such as pedestrians and cars.
    Most existing detectors are unable to detect uncommon objects and corner cases (\eg, a dog crossing a street), which may lead to severe accidents in some situations, making the timeline for the real-world application of reliable autonomous driving uncertain. 
    One main reason that impedes the development of truly reliably self-driving systems is the lack of public datasets for evaluating the performance of object detectors on corner cases.
    Hence, we introduce a challenging dataset named CODA that exposes this critical problem of vision-based detectors.
    The dataset consists of 1500 carefully selected real-world driving scenes, each containing four object-level corner cases (on average), spanning more than 30 object categories.
    On CODA, the performance of standard object detectors trained on large-scale autonomous driving datasets significantly drops to no more than 12.8\% in mAR.
    Moreover, we experiment with the state-of-the-art open-world object detector and find that it also fails to reliably identify the novel objects in CODA, suggesting that a robust perception system for autonomous driving is probably still far from reach.
    We expect our CODA dataset to facilitate further research in reliable detection for real-world autonomous driving.
    Our dataset is available at \url{https://coda-dataset.github.io}.
\end{abstract}


\definecolor{background_color}{RGB}{245, 245, 245}

\begin{figure}[t]
    \centering
    \includegraphics[width=0.85\columnwidth]{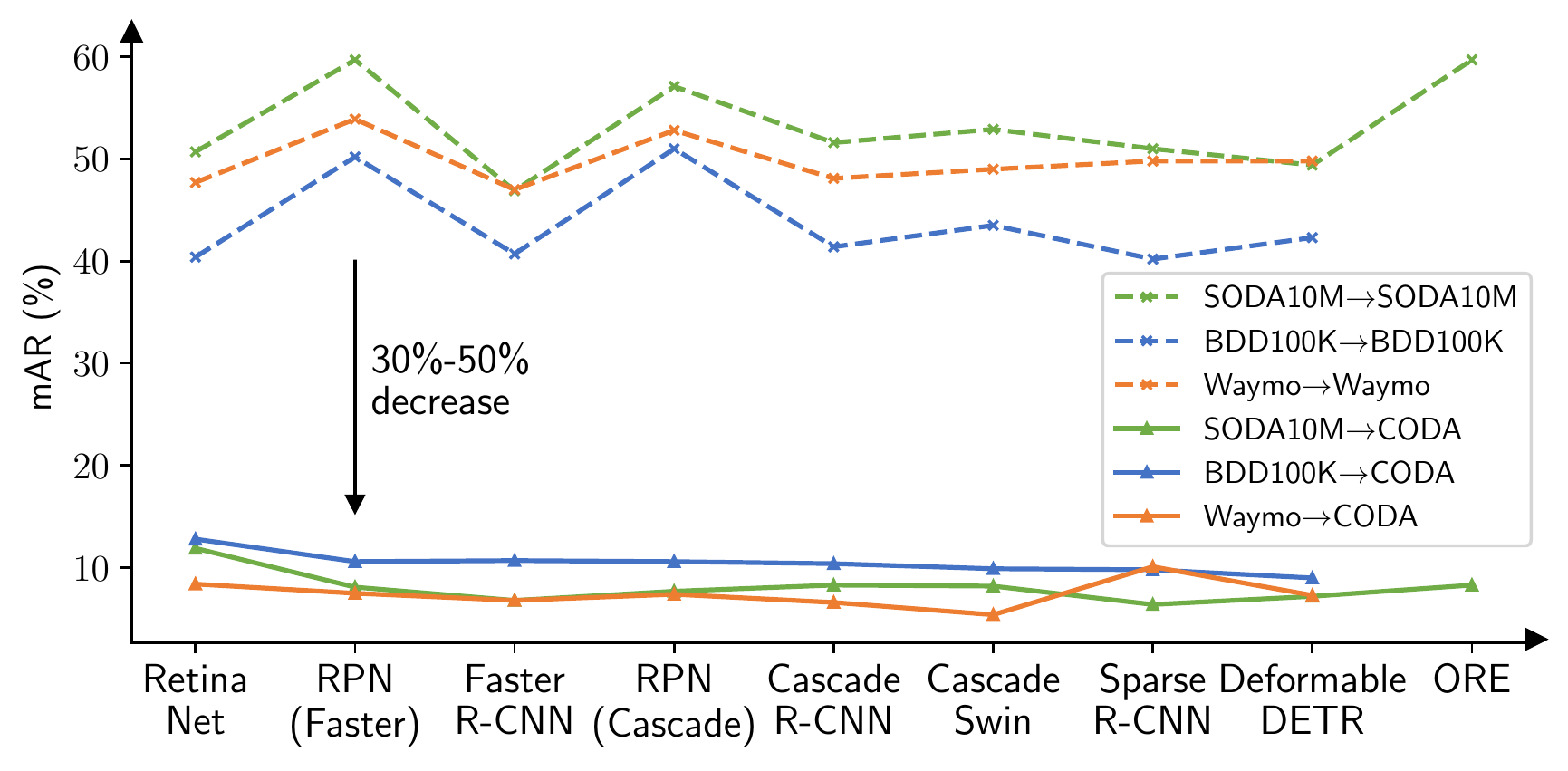}
    
    \caption{Detection results on CODA compared with common autonomous driving datasets.
    All detectors suffer from a significant 30\%-50\% performance drop, with the best achieved at 12.8\% mAR, which is definitely far from solved.
    Here \emph{A}$\rightarrow$\emph{B} represents that the detector is trained on dataset \emph{A} and evaluated on dataset \emph{B}.}
    \label{fig:performance-chart}
    
\end{figure}


\section{Introduction}
\label{sec:intro}

Deep learning has achieved prominent success in object detection for autonomous driving in the wild \cite{cai18cascadercnn,detection_waymo_spnas,sun2021sparse,zhu2020deformable}.
The success is mainly attributed to deep neural networks trained on an extensive amount of data extracted from real-life driving scenarios, which have become an indispensable component of existing autonomous driving systems~\cite{chen2021multisiam,han2021soda10m,mao2021one}.
Though such models are proficient in detecting common traffic participants (\eg, cars, pedestrians, and cyclists), they are generally incapable of detecting novel objects that are not seen or rarely seen in the training process, i.e., the out-of-distribution samples~\cite{ye2022ood,zhou22model,zhou22sparse}.
For instance, a vehicle equipped with state-of-the-art detectors galloping on the highway may fail to detect a runaway tire or an overturned truck straight ahead of the road.
These failure cases of object detection in autonomous driving may result in severe consequences, putting lives at risk.

\begin{figure}[t]
    \centering
    \begin{minipage}[t]{.193\linewidth}
        \includegraphics[width=\linewidth]{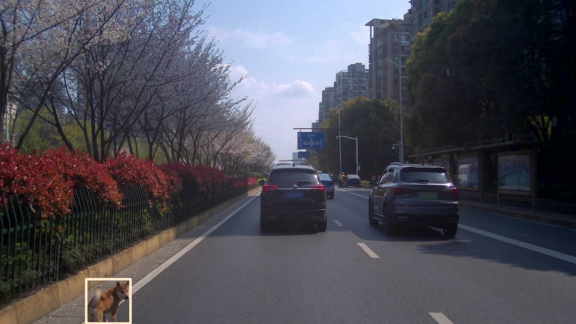}
        \vskip 1pt
        \includegraphics[width=\linewidth]{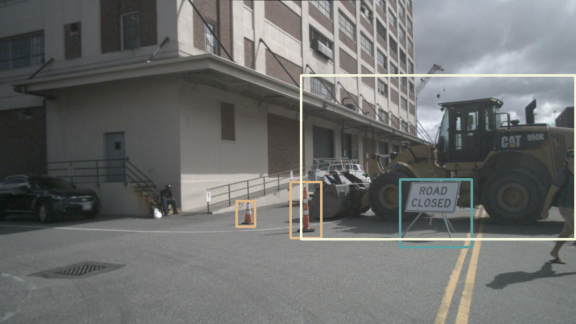}
        \vskip 1pt
        \includegraphics[width=\linewidth]{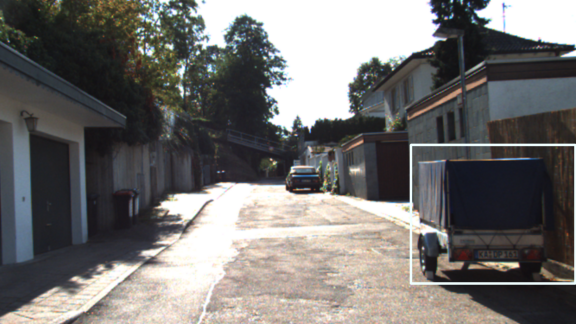}
    \end{minipage}
    \hfill
    \begin{minipage}[t]{.193\linewidth}
        \includegraphics[width=\linewidth]{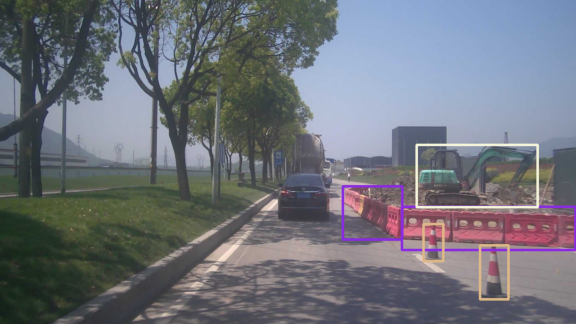}
        \vskip 1pt
        \includegraphics[width=\linewidth]{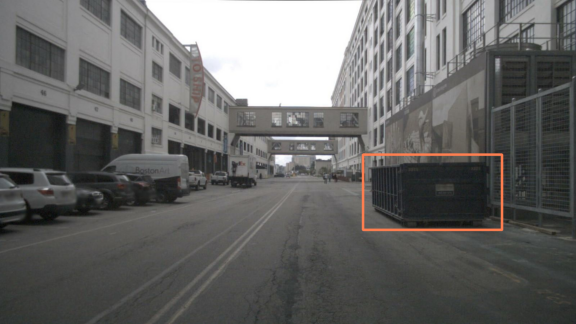}
        \vskip 1pt
        \includegraphics[width=\linewidth]{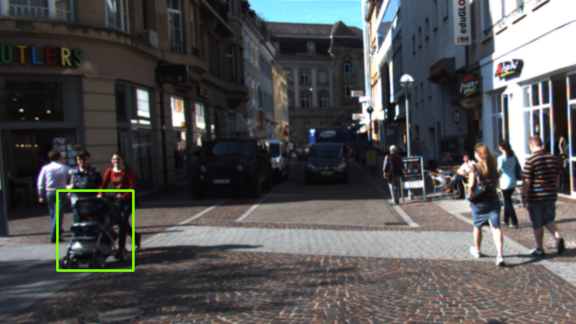}
    \end{minipage}
    \hfill
    \begin{minipage}[t]{.193\linewidth}
        \includegraphics[width=\linewidth]{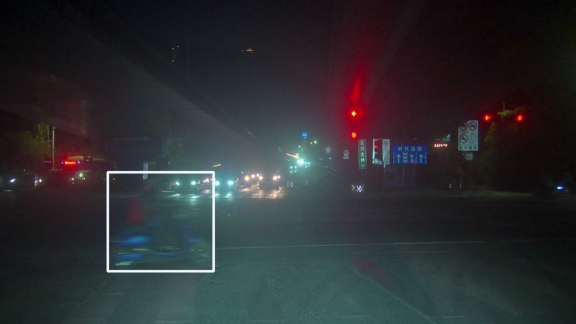}
        \vskip 1pt
        \includegraphics[width=\linewidth]{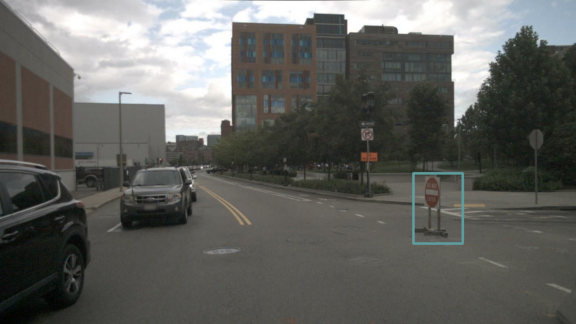}
        \vskip 1pt
        \includegraphics[width=\linewidth]{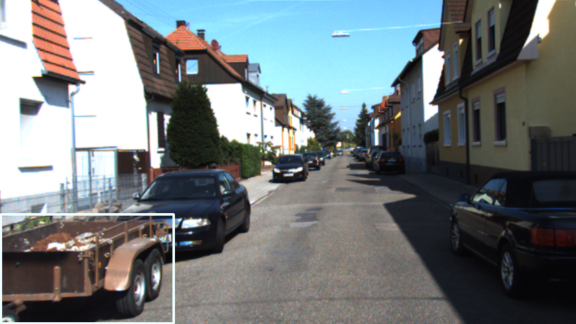}
    \end{minipage}
    \hfill
    \begin{minipage}[t]{.193\linewidth}
        \includegraphics[width=\linewidth]{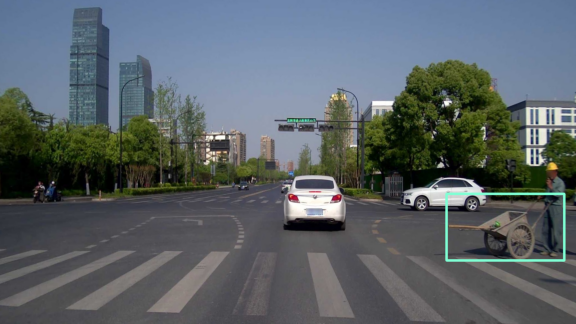}
        \vskip 1pt
        \includegraphics[width=\linewidth]{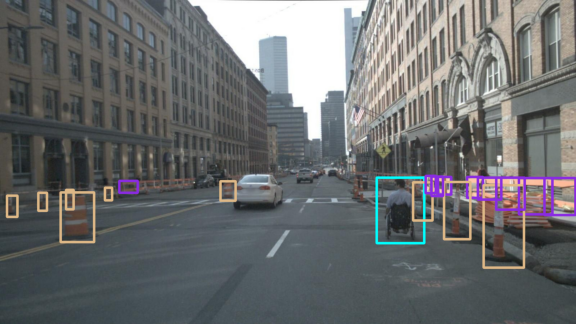}
        \vskip 1pt
        \includegraphics[width=\linewidth]{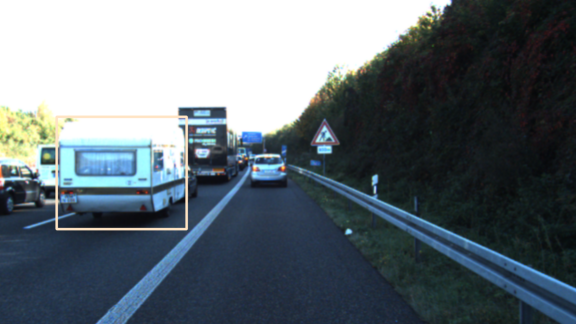}
    \end{minipage}
    \hfill
    \begin{minipage}[t]{.193\linewidth}
        \includegraphics[width=\linewidth]{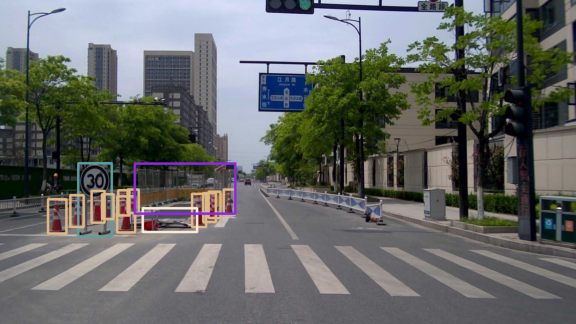}
        \vskip 1pt
        \includegraphics[width=\linewidth]{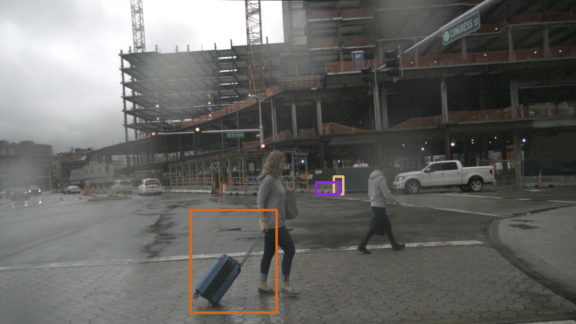}
        \vskip 1pt
        \includegraphics[width=\linewidth]{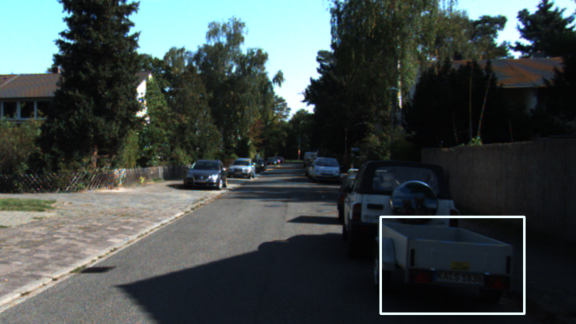}
    \end{minipage}
    
    \caption{Examples from CODA. Corner cases are indicated by the bounding boxes, while each color stands for a different object class. CODA contains both \emph{instances of novel classes} (e.g., the dog in the top-left image) and \emph{novel instances of common classes} (e.g., the cyclist in the top-middle image).}
    \label{fig:dataset-examples}
    
\end{figure}

To address the problem, we introduce CODA, a novel dataset of \emph{object-level corner cases}\footnote{We adopt the definition of object-level corner case proposed in~\cite{breitenstein2021corner}.} in real-world driving scenes.
CODA is constructed from three major object detection benchmarks for autonomous driving---KITTI~\cite{Geiger2012CVPR}, nuScenes~\cite{nuscenes2019}, and ONCE~\cite{mao2021one}.
In \cref{fig:dataset-examples}, the examples from CODA exhibit a diverse set of scenes and a great variety of novel objects.
In total, 1500 scenes (images) are selected from the combined dataset of over one million scenes, leading to nearly 6000 high-quality annotated road corner cases.
The selection process of CODA consists of two stages: a fully-automated generation of proposals on potential corner cases followed by manual inspections and corrections on the proposals.
Our approach for corner-case proposal generation, COPG, which significantly reduces the amount of human labor in the second stage, is a generic pipeline that only requires raw sensory data from camera and lidar sensor, \ie, no annotation is needed.
We believe that the approach can be utilized to efficiently produce more corner case datasets in the future.

On CODA, we have evaluated various kinds of object detection methods including standard (closed-world) detectors such as Faster R-CNN~\cite{Ren2015Faster}; a recently-proposed open-world detector, ORE~\cite{joseph2021towards}, which is capable of detecting certain objects of unseen classes; and two anomaly detection methods~\cite{memorybankanomaly,synthesize} which are also in some sense suited to the task.
Our experiment results show that none of the methods can consistently detect the novel objects in CODA, demonstrating how challenging CODA is.
In general, there is no clear winner among the methods, even though ORE shows some improvements over the closed-world detectors.
Finally, we hope that CODA can serve as an effective means for evaluating the robustness of machine perception in autonomous driving, and in turn, facilitate the development of truly reliably self-driving systems.
The main contribution of this work can be summarized as follows:
\begin{itemize}
    \item We propose CODA, the first real-world road corner case dataset, serving as a benchmark for the development of fully reliable self-driving vehicles.
    \item We evaluate various state-of-the-art object detectors (\eg, Cascade R-CNN~\cite{cai18cascadercnn}, Deformable DETR~\cite{zhu2020deformable}, and Sparse R-CNN~\cite{sun2021sparse}), suggesting that truly reliably self-driving systems are probably still far from reach.
    \item We introduce COPG, a generic pipeline for corner-case discovery, reducing human labeling effort by nearly 90\% on a large-scale dataset.
\end{itemize}


\section{Related Work}

\subsubsection{Road anomaly and corner case dataset.}
One of the pioneering datasets in road anomaly and corner case detection is the Lost and Found dataset~\cite{pinggera2016lost} which features small objects in artificial scenes.
Later introduced datasets mainly focus on semantic segmentation.
Notable ones include the road anomaly dataset of Lis~\etal~\cite{lis2019detecting} containing 60 real-world scenes, and Fishyscapes~\cite{blum2019fishyscapes}, a synthesized dataset created by overlaying objects crawled from the web onto the scenes of Cityscapes~\cite{cityscapes_dataset} and the Lost and Found dataset.
StreetHazards~\cite{hendrycks2019benchmark} is another synthesized dataset where the scenes are simulated by computer graphics.
In the same paper, the authors also introduced BDD-Anomaly, a subset of BDD100K~\cite{bdd100k}, treating trains and motorcycles as anomalous objects.


\subsubsection{Object detection.} 
Existing methods can be generally categorized into one-stage and two-stage based on how the proposals are generated.
One-stage detectors~\cite{Lin2017Focal,liu2016ssd,Redmon_2016_CVPR} densely predict class distributions and box coordinates on each position of a given image,
while two-stage detectors~\cite{cai18cascadercnn,lin2017feature,Ren2015Faster} utilize the Region Proposal Network (RPN) to generate regions of interest (RoI), which are then fed into multi-head networks for class and coordinate offset prediction.
Cascade R-CNN~\cite{cai18cascadercnn} further improves by adding a sequence of heads trained with increasing IoU thresholds.
ImageNet-supervised pre-training is adopd to accelerate training, while self-supervised pre-training~\cite{chen2021multisiam,he2020momentum,liu2022task} has recently demonstrated better transfer performance.
Previous detectors are mostly trained in the closed-world setting, which can only detect objects belonging to a pre-defined semantic class set.
To build a real-world perception system, open-world detection~\cite{joseph2021towards} has raised more attention, which can explicitly detect objects of unseen classes as \emph{unknown}.


\begin{figure}[t]
    \centering
    \includegraphics[width=\linewidth]{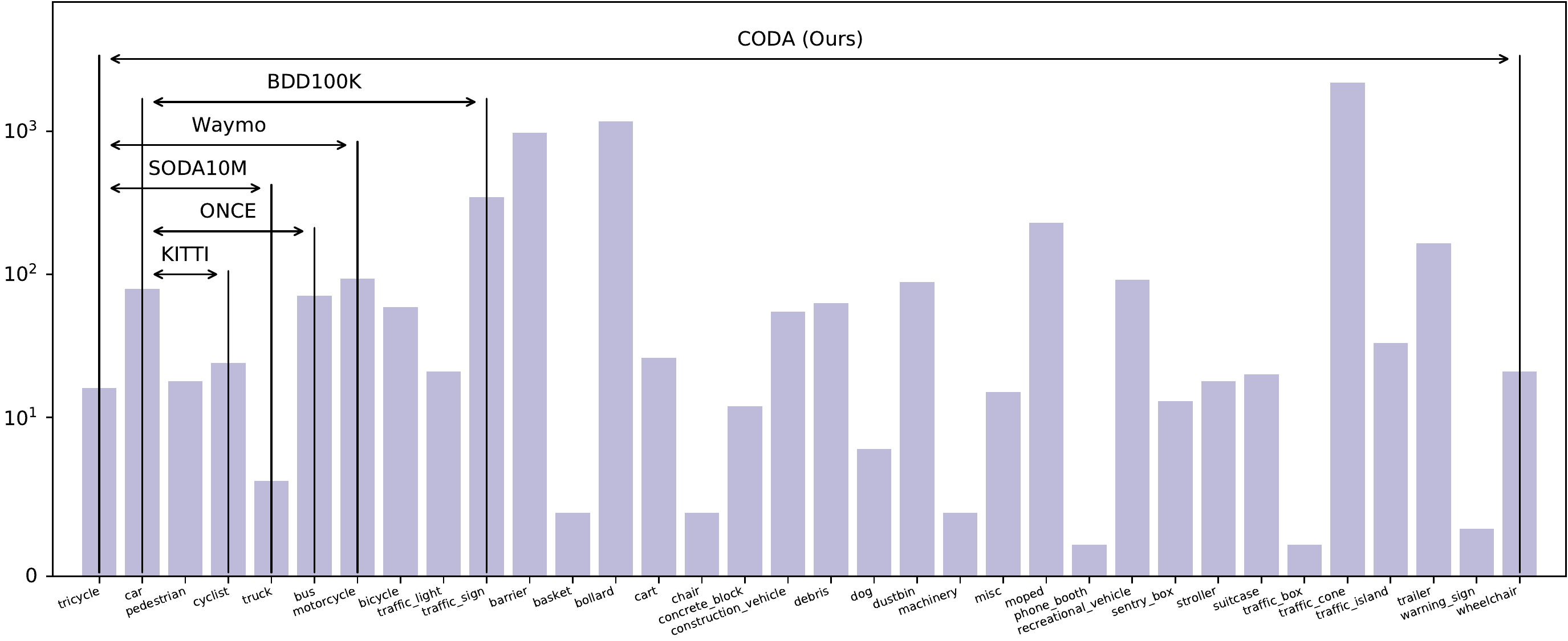}
    \caption{Class distribution of CODA and annotation coverage of common large-scale autonomous driving benchmarks in comparison with ours.
    The distribution is inherently long-tailed as suggested by Zipf's law.
    Class \emph{tram} in SODA10M and class \emph{train} in BDD100K are omitted because CODA does not contain such instance.}
    \label{fig:class-distribution}
    
\end{figure}


\section{Properties of CODA}

\subsubsection{Composition.}
The scenes in CODA are carefully selected from three large-scale autonomous driving datasets: KITTI~\cite{Geiger2012CVPR}, nuScenes~\cite{nuscenes2019}, and ONCE~\cite{mao2021one}.
Together, they contribute 1500 diverse scenes to CODA, each containing at least one object-level corner case that is hazardous to self-driving vehicles or their surrounding lives and assets.
The corner cases can be generally grouped into 7 super-classes: \emph{vehicle}, \emph{pedestrian}, \emph{cyclist}, \emph{animal}, \emph{traffic facility}, \emph{obstruction}, and \emph{misc}, governing the 34 fine-grained classes listed in \cref{fig:class-distribution}.
Moreover, these classes can be divided into \emph{novel classes} and \emph{common classes}.
Common classes stand for common object categories (e.g., cars and pedestrians) of existing autonomous driving benchmarks; whereas novel classes stand for the opposites, such as dogs and strollers.
More than 90\% of the instances in CODA are of novel classes.
On one hand, instances of novel classes are inherently undetectable by (closed-world) object detectors that are trained on the common classes.
On the other hand, the detectors ought to correctly identify novel instances of common classes, but often fail in doing so.
Detailed definitions of common/novel classes will be introduced in \cref{sec:experiment}, which is important to the evaluation of prevalent object detectors.


\subsubsection{Diversity.}

\begin{table}[t]
    \caption{Comparison with other datasets. CODA is the largest dataset of its kind in multiple aspects.
    Here we do not compare with the Fishyscapes Web dataset~\cite{blum2019fishyscapes}, which is neither publicly available nor with detailed statistics. ``$^\dagger$'' means rough estimates.}
    \label{tab:dataset-comp}
    \setlength\tabcolsep{3pt}
    \centering
    \scalebox{0.85}{
    \begin{tabular}{l|cccccc}
    \toprule
    Dataset & \#Scenes & Real & Weather & Period & \#Classes & \#Instances \\
    \midrule
    Lis~\etal~\cite{lis2019detecting} & 60 & \cmark & \xmark & \xmark & 2 & \hphantom{$^\dagger$}300$^\dagger$ \\
    Fishyscapes L\&F~\cite{blum2019fishyscapes} & 375 & \cmark & \xmark & \xmark & 3 & \hphantom{$^\dagger$}500$^\dagger$ \\
    Fishyscapes Static~\cite{blum2019fishyscapes} & 1030 & \xmark & \xmark & \xmark & 3 & \hphantom{$^\dagger$}1200$^\dagger$ \\
    StreetHazards~\cite{hendrycks2019benchmark} & \textbf{1500} & \xmark & \xmark & \xmark & 1 & \hphantom{$^\dagger$}1500$^\dagger$ \\
    BDD-Anomaly (v1)~\cite{hendrycks2019benchmark} & 361 & \cmark & \xmark & \xmark & 2 & 4476 \\
    \midrule
    CODA-KITTI (Ours) & 309 & \cmark & \cmark & \cmark & 6 & 399 \\
    CODA-nuScenes (Ours) & 134 & \cmark & \cmark & \cmark & 17 & 1125 \\
    CODA-ONCE (Ours) & 1057 & \cmark & \cmark & \cmark & 32 & 4413 \\
    \midrule
    CODA (Ours) & \textbf{1500} & \cmark & \cmark & \cmark & \textbf{34} & \textbf{5937} \\
    \bottomrule
    \end{tabular}}
\end{table}


\begin{wrapfigure}[17]{r}{4cm}
    \centering
    \includegraphics[trim={0 4cm 0 3cm}, width=\linewidth]{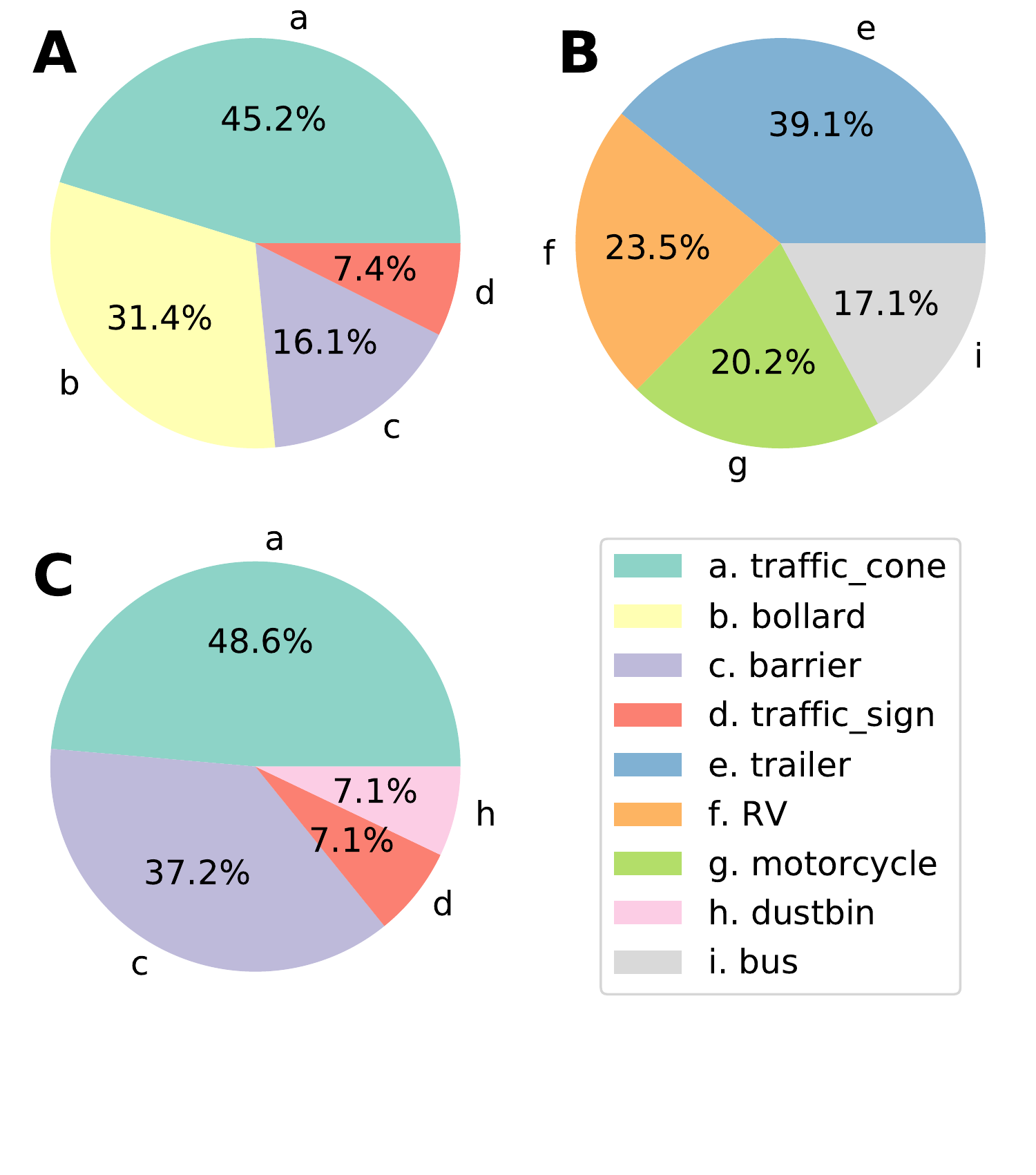}
    \caption{Distribution of the top-4 classes in the three domains of CODA: \textbf{A} ONCE, \textbf{B} KITTI, and \textbf{C} nuScenes. The distribution largely differs across the domains.}
    \label{fig:domain-distribution}
\end{wrapfigure}


The data diversity of CODA can be seen from both object level and scene level. On the object level, CODA comprises a wide range of object classes, most of which are neglected by the existing benchmarks (see \cref{fig:class-distribution}).
Though some class only has several instances (due to the natural scarcity of corner cases), they constitute a nontrivial portion of real-world driving environments.
Notably, traffic facilities such as \emph{traffic cone} and \emph{barrier} take up a majority of the corner cases because they are indeed more common and often appear in large quantities.

On the scene level, CODA contains scenes from three different countries\footnote{KITTI are captured in a mid-size city of Germany, nuScenes are captured in Singapore, and ONCE are captured in various cities of China.}, which are distinct from one another as shown by the examples in \cref{fig:dataset-examples}.
As a result, they introduce more novelty to the corner cases as the difference in object appearance is also a part of the domain shift of the scenes.
The disparity between the domains can be seen from \cref{fig:domain-distribution}, where the distribution of top-4 common classes largely differs.
In addition, the scenes in CODA exhibit different weather conditions, of which 75\% are clear, 22\% are cloudy, and 4\% are rainy. Lastly, 9\% of the scenes are night scenes apart from the daytime scenes.


\begin{figure}[t]
    \centering
    \definecolor{background_color}{RGB}{245, 245, 245}
    \definecolor{fill_color}{RGB}{255, 255, 255}
    \definecolor{draw_color}{RGB}{0, 0, 0}
    \begin{tikzpicture}[]
        \tikzstyle{operation}=[draw=draw_color, fill=fill_color, shape=rectangle, inner sep=1.5mm, rounded corners, line width=0.15mm, align=center, scale=0.6]
        \tikzstyle{image}=[inner sep=0.8mm, scale=0.9]
        \tikzstyle{mydraw}=[draw=draw_color, line width=0.15mm]
        
        \node [style=image]%
            (range_img_with_ground) {\includegraphics[width=.13\textwidth]{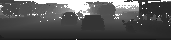}};
        
        \node [style=operation, below=2 mm of range_img_with_ground]%
            (ground_removal) {Ground-point\\removal};
        
		\node [style=image, below=2 mm of ground_removal]
		    (range_img) {\includegraphics[width=.13\textwidth]{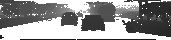}};
		
		\node [style=operation, below=2 mm of range_img]%
		    (clustering) {Point-cloud\\clustering};
		
		\node [style=image, below=2 mm of clustering, label={[label distance=1mm, align=center, scale=0.6]below:\textbf{(a) Range images}\\\textbf{\ \ \& point clusters}}]%
		    (clustered) {\includegraphics[width=.13\textwidth]{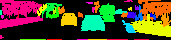}};
		
		\node [style=operation, right=5 mm of clustered]%
		    (gen_prop) {Cluster\\projection};
		
		\node [style=image, right=2 mm of gen_prop, label={[scale=0.6]below:\textbf{(e) Initial proposal}}]%
		    (init_prop) {\includegraphics[width=.13\textwidth]{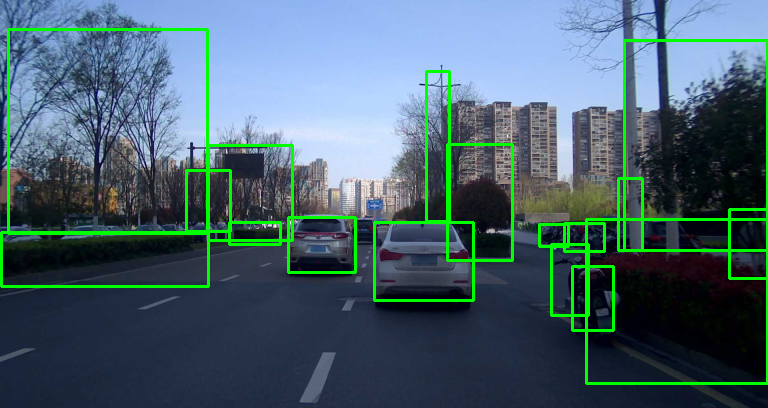}};
		
		\node [style=operation, right=2 mm of init_prop]%
		    (back_removal) {Background\\removal\vphantom{p}};
		
		\node [style=image, right=2 mm of back_removal, label={[scale=0.6]below:\textbf{(f) Intermediate proposal}}]%
		    (inter_prop) {\includegraphics[width=.13\textwidth]{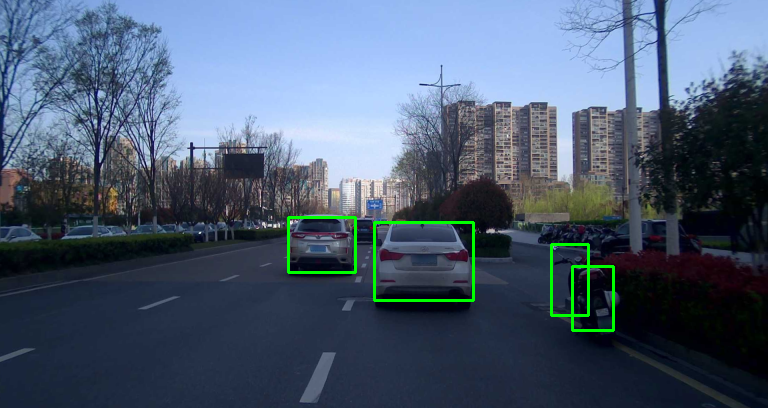}};
		    
		\node [style=operation, right=2 mm of inter_prop]%
		    (suppression) {Common-class\\suppression};
		    
		\node [style=image, right=2 mm of suppression, label={[name=label_a, scale=0.6]below:\textbf{(g) Final proposal}}]%
		    (final_prop) {\includegraphics[width=.13\textwidth]{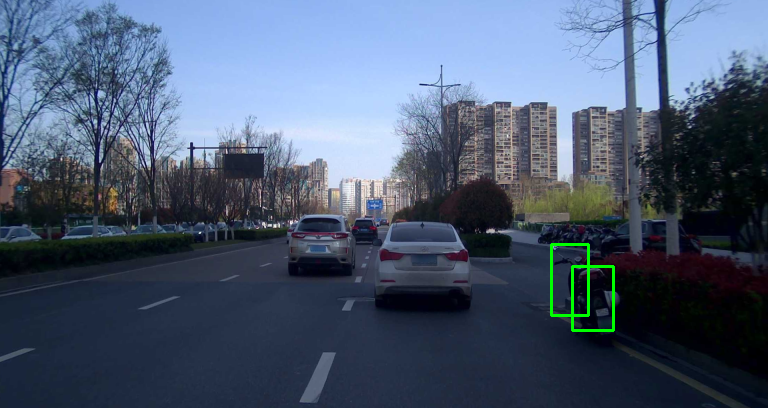}};
		    
		\node [style=image, above=7mm of gen_prop, label={[scale=0.6]below:\textbf{(b) Camera image}}]%
		    (camera_img) {\includegraphics[width=.13\textwidth]{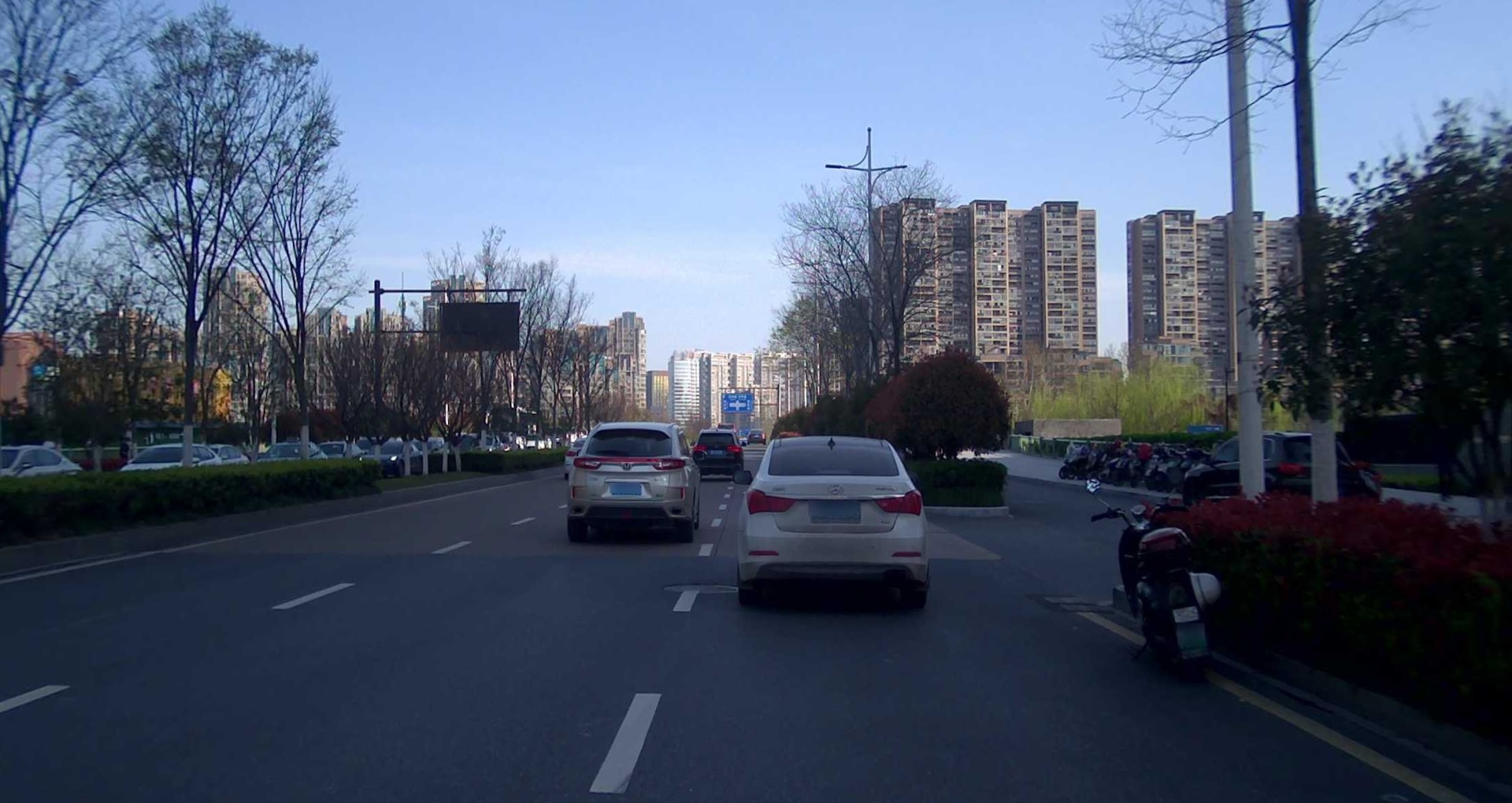}};
		    
		\node [style=image, above=7 mm of back_removal, label={[scale=0.6]below:\textbf{(c) Segmentation map}}]%
		    (seg_img) {\includegraphics[width=.13\textwidth]{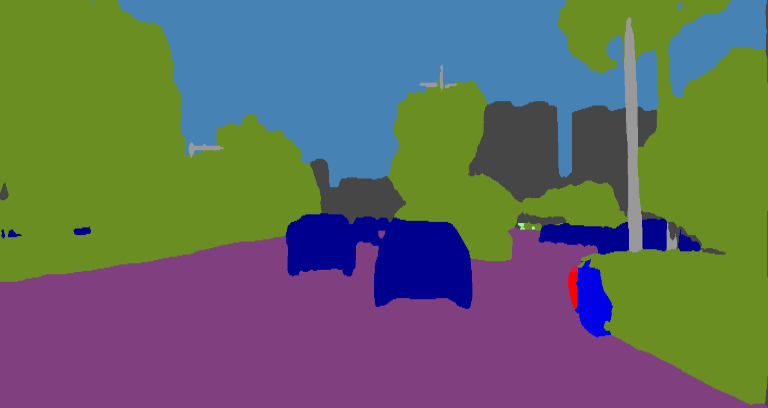}};
		    
		\node [style=image, above=7 mm of suppression, label={[scale=0.6]below:\textbf{(d) Detection result}}]%
		    (detect_result) {\includegraphics[width=.13\textwidth]{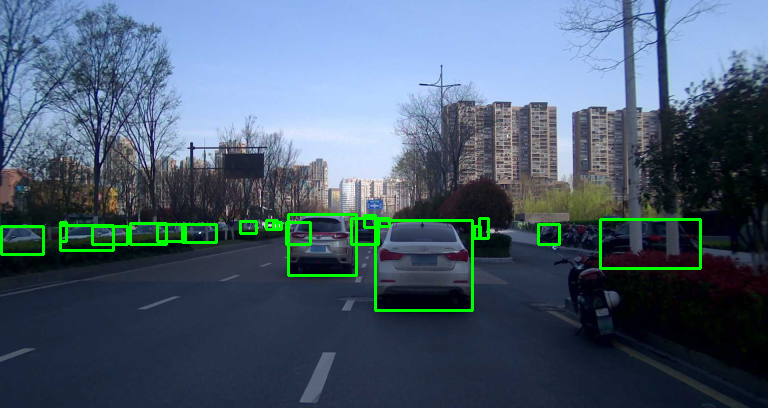}};
		    
		\node [style=operation, above=3.5 mm of seg_img]%
		    (segmentation) {Semantic segmentation};
		    
		\node [style=operation, above=3.5 mm of detect_result]%
		    (detection) {Object detection};
		    
		\draw [-       , style=mydraw] (range_img_with_ground) -- (ground_removal);
		\draw [-stealth, style=mydraw] (ground_removal) -- (range_img);
		\draw [-       , style=mydraw] (range_img) -- (clustering);
		\draw [-stealth, style=mydraw] (clustering) -- (clustered);
		\draw [-       , style=mydraw] (clustered) -- (gen_prop);
		\draw [-stealth, style=mydraw] (gen_prop) -- (init_prop);
		\draw [-       , style=mydraw] (init_prop) -- (back_removal);
		\draw [-stealth, style=mydraw] (back_removal) -- (inter_prop);
		\draw [-       , style=mydraw] (inter_prop) -- (suppression);
		\draw [-stealth, style=mydraw] (suppression) -- (final_prop);
		\draw [-       , style=mydraw] ($(camera_img.south) - (0, 4mm)$) -- ($(gen_prop.north)$);
		\draw [-       , style=mydraw] (camera_img) |- ($(segmentation.north) + (0, 3.5mm)$) -- (segmentation);
		\draw [-       , style=mydraw] ($(segmentation.north) + (0, 3.5mm)$) -| (detection);
		\draw [-stealth, style=mydraw] (segmentation) -- (seg_img);
		\draw [-       , style=mydraw] ($(seg_img.south) - (0, 4mm)$) -- (back_removal);
		\draw [-stealth, style=mydraw] (detection) -- (detect_result);
		\draw [-       , style=mydraw] ($(detect_result.south) - (0, 4mm)$) -- (suppression);
		\begin{scope}[on background layer]
            \node (b0) [fill=background_color, inner sep=0, fit margins={left=2mm,right=1mm,bottom=1.5mm,top=1.5mm}, rounded corners,  fit=(range_img_with_ground)(clustered)(label_a)(final_prop)] {};
        \end{scope}
    \end{tikzpicture}
    
    \caption{Pipeline for generating proposals of corner case (COPG). The input to the pipeline is the point cloud and the camera image of a given scene. The point cloud is used to compute \textbf{(a)}, whereas the camera image \textbf{(b)} is used to produce \textbf{(c)} and \textbf{(d)}, which then help remove invalid proposals. The output \textbf{(g)} is a set of bounding boxes indicating the proposed corner cases in the camera image.}
    \label{fig:proposal-pipeline}
\end{figure}
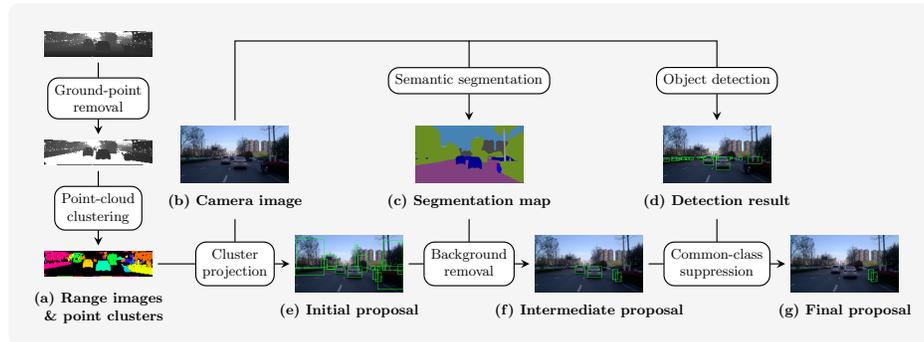


\subsubsection{Comparison with road anomaly datasets.}
In \cref{tab:dataset-comp}, we compare CODA against several prominent road anomaly datasets that also have object-level annotations.
In contrast to CODA, the datasets are either synthetic or small in scale.
The largest one of real-world road anomalies, BDD-Anomaly (v1)~\cite{hendrycks2019benchmark} only contains two object classes, albeit it is comparable to CODA on the number of instances.


\section{Construction of CODA}
As mentioned earlier, CODA is constructed from three autonomous driving benchmarks, of which most scenes are captured in well-regulated urban areas and therefore contain very few corner cases.
To identify them in the large pools of data, we must first define what ``corner cases'' are in a clearer sense.
The main criteria we use for determining whether an object is a corner case are as follows:
\begin{itemize}
    \setlength\itemsep{0.3em}
    \item \textbf{Risk:} The object blocks or is about to block a potential path of the self-driving vehicle mounted with the camera. Static objects not on the road such as trees and buildings are not considered to block the vehicle.
    \item \textbf{Novelty:} The object does not belong to any of the common classes of autonomous driving benchmarks, or it is a novel instance of the common classes. For simplicity, we take the classes of SODA10M~\cite{han2021soda10m} as the common classes.
\end{itemize}
If an object satisfies both criteria then it is a corner case.
The first criterion suggests that the object could be hit by the vehicle and the second criterion suggests that the object is difficult to detect.

\subsection{Overview}
Adhering to the high-level criteria above, the construction of CODA is carried out in two main stages. The first stage is an automatic generation of proposals that identifies potential corner cases from initial data, followed by the second stage, a manual selection and labeling process that eliminates the false positives of the proposals, and then classifies the remaining true positives while adjusting their bounding boxes to be more precise.

For ONCE~\cite{mao2021one} consisting of a million scenes, the first stage helps filter out nearly 90\% of scenes that are unlikely to contain any corner case, significantly reducing human efforts in the subsequent stage.
For KITTI~\cite{Geiger2012CVPR} and nuScenes~\cite{nuscenes2019}, which are considerably smaller than ONCE, we skip the first stage by adopting the ground-truth annotations of uncommon objects that are already provided by the datasets as proposals.

Next, we introduce COPG, our pipeline for corner-case proposal generation (illustrated in \cref{fig:proposal-pipeline}). It only requires raw sensory data from a camera and a lidar sensor, \ie, 2D images and 3D point clouds, to identify potential corner cases in any given dataset.


\subsection{Identifying Potential Corner Cases}\label{sec:COPG}

\begin{wrapfigure}[24]{r}{5cm}
    \centering
    \includegraphics[trim={0 1.5cm 0 3.5cm}, width=\linewidth]{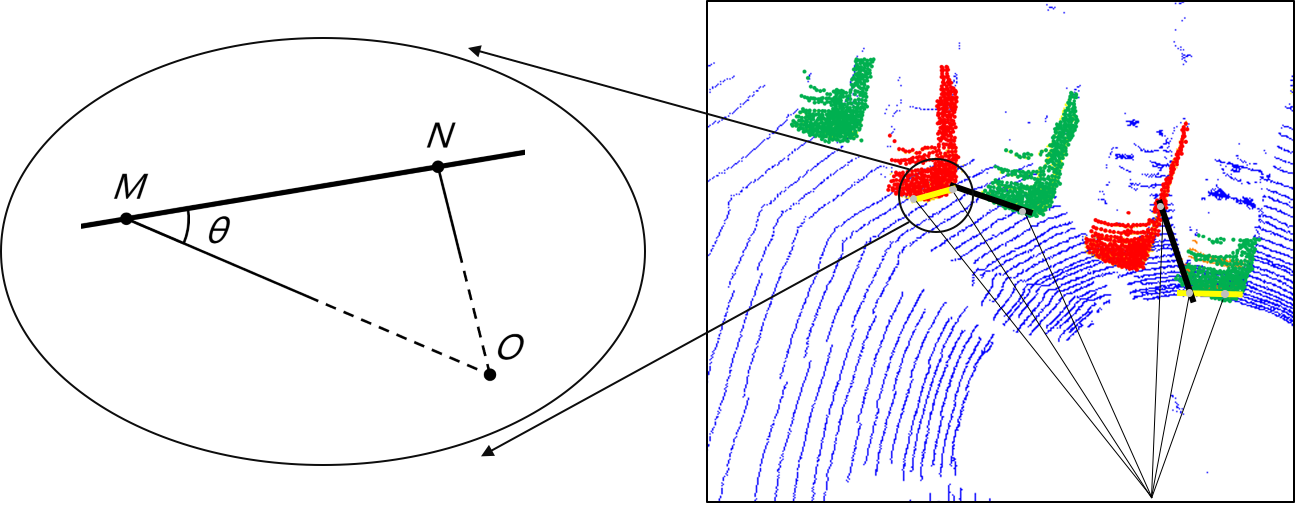}
    \caption{Abstraction of the point-cloud clustering algorithm~\cite{bogoslavskyi2016fast}.
    The right figure is a top-view example separating five cars.
    In the left figure, $O$ denotes the location of the lidar sensor, $M$ and $N$ denote two points in the cloud, while $OM$ and $ON$ denote two lidar beams ($OM$ is the longer beam). If the angle $\theta$ is greater than a fixed threshold, then the algorithm labels $M$ and $N$ as points belonging to the same object.
    The rule is based on the observation that in most cases, if $M$ and $N$ are from the same object, $\theta$ is relatively large; however, for those from different objects, $\theta$ turns out to be substantially smaller.}
    \label{fig:clustering}
\end{wrapfigure}

\subsubsection{Unsupervised point-cloud clustering.}
To reliably identify objects satisfying the first criterion, the first step is to learn the location of nearby objects that could obstruct the road. Hence, we turn to lidar point clouds.
Since we do not assume any annotation on the points, we start by clustering them so as to separate the objects in the cloud.
But before that, we remove all ground-level points by RANSAC~\cite{fischler1981random} to avoid ground points being then clustered as parts of other objects and to suppress the noise from insignificant objects (\eg, tin cans and small branches) on the ground.

Given a point cloud with ground-level points removed, we adopt the algorithm proposed by Bogoslavskyi and Stachniss~\cite{bogoslavskyi2016fast} to cluster the remaining points.
The algorithm operates on the range image of the point cloud.
A range image is a 2D image showing the distance to points in a scene from a specific point (which is the location of the lidar sensor in our case) and the image has pixel values that correspond to the distance.
Given a range image, the algorithm conducts a breadth-first search over the pixels of the image, and eventually assigns every pixel to a cluster.
Specifically, the algorithm compares each pixel $p$ with its four neighboring pixels during the search. If a neighbor $p^\prime$ is sufficiently \emph{close} to $p$, then they are given the same cluster label.
The closeness between pixels is determined by the geometric relationship between the underlying points (in the 3D cloud) of these pixels.
See \cref{fig:clustering} for a detailed explanation.

After separating points of different objects apart by the clustering algorithm, the points are projected onto the camera images.
2D bounding boxes are then generated for each of the clusters, except those that are too small or too far away from the lidar sensor.
These bounding boxes are our initial proposals of corner cases, which is a superset of the final proposals.
Next, we apply two other techniques to remove the proposals that violate the predefined criteria.


\subsubsection{Background removal.}
Not all objects found by the point-cloud clustering algorithm satisfy the first criterion since most of the objects are usually off the road.
Static objects in the background (\eg, vegetation and buildings) are the most common ones in this category.
Discerning these objects from the others requires a semantic understanding of the scene, which could not be derived from merely point-cloud data.
Instead, we find semantic segmentation on camera images particularly useful.
We utilize a DeepLabv3+~\cite{deeplabv3plus2018} model pre-trained on Cityscapes~\cite{cityscapes_dataset} to produce fine segmentation maps, and then filter out the proposals that has a large overlap (over some threshold) with background regions in the corresponding segmentation map. The following classes are considered as backgrounds: \emph{road}, \emph{sidewalk}, \emph{building}, \emph{wall}, \emph{fence}, \emph{pole}, \emph{vegetation}, \emph{terrain}, and \emph{sky}.
After removing the backgrounds, we obtain a set of objects that mostly agrees with our first criterion.

\subsubsection{Common-class suppression.}
To meet the second criterion, the one on the novelty of corner case, we make use of object detectors used in autonomous driving systems to filter out objects that are not considered novel by our standard.
Specifically, we utilize Cascade R-CNN~\cite{cai18cascadercnn} with SP-Net backbone~\cite{detection_waymo_spnas} trained on a private dataset that is similar to ONCE and consists of millions of scenes to detect common-class objects, producing a set of bounding boxes for each scene.
The bounding boxes are subsequently compared with the proposals from the previous step, and those proposals that have IoUs over a threshold with any of the detected common objects are removed.

Note that we do not use ground-truth annotations to suppress the common-class proposals.
Our approach has two important advantages: 1) it applies to unlabeled data as long as there is a working detector trained on a similar dataset of the same task; and 2) it keeps some novel instances of the common classes, \ie the ``hard cases'' in object detection, that would otherwise be suppressed by the ground truth.
The effectiveness of COPG is demonstrated in \cref{sec:effectiveness-COPG}.

\subsection{Further Examination}
\label{sec:manual_select}
In the previous subsection, we have discussed how to extract potential corner cases from an abundance of unlabeled data.
On ONCE~\cite{mao2021one}, the process leaves only around 10\% of the scenes for further examination.
It is perhaps worth noting that by increasing the thresholds of background removal and common-class suppression, one can further reduce the number of candidate scenes, but it would also cause more corner cases to be neglected.
In some sense, the thresholds control the trade-off between the final amount of true positives and the amount of human labor required to pick them out (see Appendix~B for relevant ablations).

\subsubsection{Selection.}
Given the generated proposals of ONCE, we start by examining the scenes containing these proposals.
Those that do not contain any valid corner case (according to our criteria) are discarded.
After the process, we finally arrive at the 1057 scenes of CODA-ONCE, roughly 0.1\% of the one million scenes in the original dataset.
This shows that corner cases are indeed rare in real-world data.
As for KITTI~\cite{Geiger2012CVPR} and nuScenes~\cite{nuscenes2019}, without undergoing the automatic generation of proposals, all data are manually selected, resulting in 309 and 134 scenes respectively.

\subsubsection{Labeling.}
To ease the labeling process, we use CLIP~\cite{radford2021learning} to pre-label the objects in CODA.
After that, we use the toolkit~\cite{labelme2016} inspired by LabelMe~\cite{russell2008labelme} to label the class of each corner case and to revise the bounding boxes since the proposals in each scene may not all be valid and the projection from point clouds to camera images is often inaccurate.
Meanwhile, some bounding boxes are also added to corner-case objects missed by the proposals in the selected scenes.
For quality assurance, the output of each annotator is verified by two other annotators.
In the end, most of the corner cases are given a label of a specific class, except the ones that are either unrecognizable or difficult to categorize, which are placed under the \emph{misc} class.


\section{Experiment} \label{sec:experiment}

\begin{table}[t]
    \centering
    \caption{Detection results (\%) on CODA.
    The best performance is achieved at 12.8\% AR, suggesting that truly reliable object detection is probably still far from reach.
    Definitions of ORIGIN, CORNER, COMMON, and NOVEL are provided in ``class separation'' of Sec.~\ref{sec:implement}.
    ``D-DETR'' is short for Deformable DETR and ``Cascade Swin'' stands for Swin-Tiny-based Cascade R-CNN.
    \textbf{Bold} values highlight the best performance among detectors pre-trained on the same dataset,
    and ``$^{\dag}$'' means official checkpoints are adopted.
    ``*'' indicates that AR is the primary evaluation metric on CODA, while 
    ``-'' suggests that the detector cannot report the corresponding values, with reasons explained in ``evaluation'' of Sec.~\ref{sec:implement}.
    See more results in Appendix~D.
    }
    
    \label{tab:exp_coda}
    \setlength\tabcolsep{1.0pt}
    \scalebox{0.76}{
    \begin{tabular}{l|c|cc|cccc|cccc|cccc}
    \toprule 
    \multicolumn{2}{c|}{CODA} & \multicolumn{2}{c|}{ORIGIN} & \multicolumn{4}{c|}{CORNER} & \multicolumn{4}{c|}{COMMON} & \multicolumn{4}{c}{NOVEL} \\
    \hline
    Method & Dataset & AP & AR & AR$^*$ & AR$_{50}$ & AR$_{75}$ & AR$^{10}$ & AR$^*$ & AR$_{50}$ & AR$_{75}$ & AR$^{10}$ & AR$^*$ & AR$_{50}$ & AR$_{75}$ & AR$^{10}$ \\
    \hline 
    RetinaNet$^{\dag}$~\cite{Lin2017Focal} & & 34.0 & 50.7 & \textbf{11.9}  & 25.2  & 9.5 & 5.4 & 28.7 & 58.9  & 23.5 & 23.9 & - & - & - & -  \\
    Faster R-CNN$^{\dag}$~\cite{Ren2015Faster} & & 36.7 & 46.9 & 6.8   & 13.0  & 6.4 & 4.9   & 23.9  & 46.8  & 20.1 & 23.1 & - & - & - & - \\
    Cascade R-CNN$^{\dag}$~\cite{cai18cascadercnn} & & 39.4 & 51.6 & 8.3   & 15.5  & 7.6  & 5.5   & 27.2  & 47.0  & 29.4 & 25.3 & - & - & - & -  \\
    D-DETR~\cite{zhu2020deformable} & & 31.8 & 49.4 & 7.2 & 16.7 & 4.9 & 3.6 & \textbf{34.6} & 60.2 & 36.5 & 29.6 & - & - & - & -  \\
    Sparse R-CNN~\cite{sun2021sparse} & SODA10M & 31.2 & 51.0 & 6.4 & 13.2 & 5.4 & 3.9 & 26.4 & 47.1 & 25.6 & 23.0 & - & - & - & -  \\
    Cascade Swin~\cite{liu2021swin} & \cite{han2021soda10m} & 41.1 & 52.9 & 
    8.2 & 15.5 & 7.6 & 5.7 & 30.4 & 51.3 & 32.2 & 29.3 & - & - & - & -  \\
    RPN (Faster)$^{\dag}$~\cite{Ren2015Faster} & & - & 59.7 & 8.1 & 16.2  & 7.4  & 3.1 & - & - & - & - & - & - & - & - \\
    RPN (Cascade)$^{\dag}$~\cite{cai18cascadercnn} & & - & 57.1 & 7.7 & 16.0  & 6.8 & 2.8 & - & - & - & - & - & - & - & -  \\
    ORE~\cite{joseph2021towards} & & \textbf{49.2} & \textbf{59.7} & 8.3   & 16.4  & 7.4   & 5.6   & 18.5  & 35.5  & 18.2 & 18.1  & \textbf{3.4}   & 7.6   & 2.8   & 2.9  \\
    \hline 
    RetinaNet$^{\dag}$~\cite{Lin2017Focal} & & 28.6 & 40.4 & \textbf{12.8} & 23.2  & 11.9  & 4.8   & 27.5  & 58.1  & 21.5  & 23.6  & \textbf{9.7} & 17.7  & 9.1   & 5.9  \\
    Faster R-CNN$^{\dag}$~\cite{Ren2015Faster} & & 31.0 & 40.7 & 10.7  & 19.2  & 10.2  & 4.3   & 24.4  & 48.1  & 20.9  & 22.0  & 7.2   & 13.3  & 6.8   & 5.9  \\
    Cascade R-CNN$^{\dag}$~\cite{cai18cascadercnn} & & 32.4 & 41.4 & 10.4  & 18.5  & 9.7   & 4.5   & 25.7  & 48.4  & 23.3 & 23.6  & 6.9 & 12.5  & 6.5 & 5.7  \\
    D-DETR~\cite{zhu2020deformable} & BDD100K & 28.5 & 42.3 & 9.0 & 22.2 & 5.6 & 2.8 & 28.5 & 63.0 & 22.3 & 26.2 & 7.0 & 17.3 & 4.3 & 3.9 \\ 
    Sparse R-CNN$^{\dag}$~\cite{sun2021sparse} & \cite{bdd100k} & 26.7 & 40.2 & 9.8 & 19.0 & 8.9 & 4.5 & 27.4 & 51.7 & 25.8 & 24.3 & 8.0 & 15.4 & 7.4 & 5.1 \\
    Cascade Swin~\cite{liu2021swin} & & \textbf{34.5} & 43.5 & 9.9 & 17.2 & 9.7 & 4.9 & \textbf{31.0} & 55.0 & 29.9 & 29.4 & 6.5 & 11.4 & 6.4 & 5.9 \\
    RPN (Faster)$^{\dag}$~\cite{Ren2015Faster} & & - & 50.2 & 10.6  & 20.0  & 10.2  & 3.7 & - & - & - & - & - & - & - & -  \\
    RPN (Cascade)$^{\dag}$~\cite{cai18cascadercnn} & & - & \textbf{51.0} & 10.6  & 20.0  & 10.2  & 3.9 & - & - & - & - & - & - & - & -  \\
    \hline 
    RetinaNet~\cite{Lin2017Focal} & & 39.7 & 47.7 & 8.4 & 15.6 & 7.7 & 5.1 & 24.5 & 43.2 & 24.4 & 22.2 & 6.7 & 11.9 & 6.4 & 4.6 \\ 
    Faster R-CNN~\cite{Ren2015Faster} & & 40.9 & 47.0 & 6.8 & 12.4 & 6.4 & 4.8 & 20.9 & 36.0 & 19.6 & 19.1 & 5.5 & 9.6 & 5.2 & 4.3  \\
    Cascade R-CNN~\cite{cai18cascadercnn} & & 42.6 & 48.1 & 6.6 & 11.4 & 6.6 & 5.0 & 18.9 & 32.6 & 20.1 & 17.6 & 5.3 & 8.7 & 5.5 & 4.4  \\
    D-DETR~\cite{zhu2020deformable} & Waymo & 40.4 & 49.8 & 7.3 & 15.8 & 5.4 & 3.6 & 28.5 & 49.4 & 24.6 & 22.5 & 5.2 & 11.5 & 4.0 & 3.0 \\ 
    Sparse R-CNN~\cite{sun2021sparse} & \cite{Sun2020Scalability} & 38.8 & 49.8 & \textbf{10.1} & 19.6 & 9.0 & 4.7 & \textbf{29.5} & 51.8 & 27.0 & 22.1 & \textbf{7.6} & 14.3 & 7.1 & 4.2 \\
    Cascade Swin~\cite{liu2021swin} & & \textbf{44.2} & 49.0 & 5.4 & 8.7 & 5.5 & 4.4 & 21.8 & 38.1 & 18.8 & 21.3 & 4.3 & 6.7 & 4.6 & 3.7 \\
    RPN (Faster)~\cite{Ren2015Faster} & & -  & \textbf{53.9} & 7.5 & 13.7 & 7.5 & 3.6 & - & - & - & - & - & - & - & -  \\
    RPN (Cascade)~\cite{cai18cascadercnn} & & - & 52.8 & 7.4 & 13.8 & 7.3 & 3.9 & - & - & - & - & - & - & - & -  \\
    \bottomrule 
    \end{tabular}}
    
\end{table}

\subsection{Implementation details} \label{sec:implement}
\subsubsection{Baselines.}
Four categories of baselines are evaluated on CODA: 
1) for \emph{closed-world object detectors}, state-of-the-art detectors of both one-stage (\eg, RetinaNet~\cite{Lin2017Focal}) and two-stage (\eg, Faster~\cite{Ren2015Faster} and Cascade R-CNN~\cite{cai18cascadercnn}) pre-trained on SODA10M~\cite{han2021soda10m}, BDD100K~\cite{bdd100k} and Waymo~\cite{Sun2020Scalability} are selected;
2) \emph{region proposal network (RPN)}~\cite{Ren2015Faster} can recognize foreground objects in a class-agnostic manner, which might learn a more generalizable representation, so we further report the performance of the RPN of Faster R-CNN and Cascade R-CNN;
3) for \emph{open-world object detectors}, we adopt the state-of-the-art ORE model~\cite{joseph2021towards} but without incremental learning; and
4) for \emph{anomaly detection}, we modify the synthesize then compare~\cite{synthesize} and memory-based OOD detection~\cite{memorybankanomaly} to generate anomaly bounding boxes based on the proposals of a pre-trained RPN.


\subsubsection{Optimization.}
We adopt ResNet-50~\cite{he2016deep} initialized with ImageNet-supervised pre-trained weights as the backbone for all baselines except Swin Transformer~\cite{liu2021swin} based Cascade R-CNN, denoted as \textit{Cascade Swin} in Tab.~\ref{tab:exp_coda}.
We utilize the officially released checkpoints of closed-world detectors pre-trained on SODA10M and BDD100K, while re-implementing all selected baselines on Waymo, whose official checkpoints are not available, using the MMDetection~\cite{mmdetection} toolbox.
All the BDD100K and Waymo baselines are trained with a batch size of 16 for 12 epochs with an 1000-iteration warmup using the SGD optimizer.
The learning rate is set as 0.02, decreased by a factor of 10 at the 8th and 11th epoch.
Lastly, we construct ORE based on Faster R-CNN using Detectron2~\cite{wu2019detectron2} following the original paper, which is then trained on SODA10M with a batch size of 8 for 24 epochs, the same with the closed-world counterparts.
More optimization details are provided in Appendix~A.


\subsubsection{Class separation.}
Considering the fact that the semantic class sets of SODA10M, BDD100K, and Waymo differ from each other, all of which are just subsets of the CODA class set, a unified separation of common and novel classes is necessary for a fair comparison of different detectors.
Without loss of generality, we define:
\textbf{1) COMMON} classes as the class set of SODA10M (\ie, \emph{pedestrian}, \emph{cyclist}, \emph{car}, \emph{truck}, \emph{tram}, and \emph{tricycle}), since ORE is trained on SODA10M;
\textbf{2) NOVEL} classes as the remaining classes of CODA beyond COMMON;
\textbf{3) CORNER} combines all COMMON and NOVEL classes to match detector predictions in a class-agnostic manner since it is more important to detect an obstacle before distinguishing its semantic class; and
\textbf{4) ORIGIN} reports detector performance on their pre-trained datasets for reference (\ie, SODA10M test set for SODA10M detectors and the corresponding validation sets for BDD100K and Waymo) since robustness to corner cases should not come at a high cost of detection precision.


\subsubsection{Evaluation.}
By the class separation described above, we divide detector predictions according to the corresponding semantic classes.
Specifically, we treat all predictions but the ORE \emph{unknown}, which should be considered as predictions for NOVEL objects, of SODA10M detectors as predictions for COMMON objects.
Predictions of \emph{pedestrian}, \emph{rider}, \emph{car}, \emph{truck}, and \emph{train} of BDD100K detectors are considered as COMMON, while the remaining ones are considered as NOVEL.
Such a disjoint division, however, is not applicable for Waymo.
According to the official document, all recognizable vehicles are annotated as \emph{vehicle} uniformly, suggesting that Waymo baselines can only detect vehicles in a class-agnostic manner.
So here, the \emph{vehicle} predictions of Waymo detectors are not only considered as COMMON (along with \emph{pedestrian} and \emph{cyclist}), but also considered as NOVEL, which might put Waymo detectors at advantage, especially for the recall-based evaluation described below, but it does not affect our conclusion.
We further project all detected COMMON vehicles to a unified \emph{vehicle} class so that detectors of different datasets have the same COMMON class set, \ie \emph{pedestrian}, \emph{cyclist}, and \emph{vehicle};
while we combine all NOVEL objects to evaluate in a class-agnostic manner since detectors cannot discriminate unseen classes.

Note that under two circumstances, detectors cannot be evaluated (marked as ``-'' in~\cref{tab:exp_coda}), including:
1) RPNs can only perform class-agnostic detection, which are only evaluated under ORIGIN and CORNER; and
2) closed-world detectors pre-trained on SODA10M cannot recognize any NOVEL objects, whose semantic class set is considered as CODA COMMON class set. 

We utilize the COCO-style Average Recall (AR) as the evaluation metrics instead of Average Precision (AP) since the annotated objects are the most challenging \textbf{subset} of all CODA foreground objects.
A model that can detect all foreground objects, including those not obstructing the road, would in fact have low AP on CODA.
Hence, AR is much more informative than AP.
We also consider
1) AR$_{50}$ and AR$_{75}$ for IoU thresholds of 0.5 and 0.75;
2) AR$^{1}$ and AR$^{10}$ for at most 1 and 10 boxes per image; and
3) AR$^{s}$, AR$^{m}$ and AR$^{l}$ for different box scales following COCO definition~\cite{coco_dataset}.


\subsection{Results}

\subsubsection{Significance of CODA.}
Experiment results are reported in \cref{tab:exp_coda}.
As summarized in \cref{fig:performance-chart}, detectors suffer from a significant performance drop of 30\%-50\% AR when deployed on CODA (\eg, 43.3\% decrease for SODA10M Cascade R-CNN).
Even for COMMON classes, the average decrease has also exceeded 21\%.
The best performance is achieved at 12.8\% AR, which is still far from solved even considering the domain gap between CODA and pre-trained datasets.
See more complete performance statistics in Appendix~D.


\subsubsection{Detectors.} 
As shown in \cref{tab:exp_coda}, Cascade R-CNN outperforms Faster R-CNN on CODA in general, not only for COMMON classes but also in the setting of CORNER class with a consistent improvement on the ORIGIN datasets, 
demonstrating the possibility to achieve higher AR on CODA without a decrease of AP on common datasets for more powerful detectors.
On the contrary, RetinaNet exceeds Cascade R-CNN at the expense of AP drop, probably due to the dense prediction design.
Note that Cascade R-CNN performs comparably or even better than RetinaNet referring to AR$^{10}$ (\eg, 5.5\% vs. 5.4\% pre-trained on SODA10M), suggesting that the AR improvement might come from more box predictions
(\eg, averaged 86 and 21 boxes/image for SODA10M RetinaNet and Cascade R-CNN).
RPN brings minor improvement but is significantly surpassed by RetinaNet even though RPN generates more box predictions (\eg, 1000 vs. 86 boxes/image on SODA10M), showing that class-aware training might be beneficial to learn a more discriminative and robust detector.
Surprisingly, we observe that ORE, the open-world detector, brings improvement on both CODA and SODA10M test set, about which more analyses are provided in \cref{sec:discussion}.


\subsubsection{Pre-train datasets.}
BDD100K detectors perform the best among three datasets, especially for the NOVEL class since BDD100K has the largest annotated semantic class set, which is definitely beneficial to detect more complicated objects and learn a more discriminative representation as previously discussed.
However, it is impossible to annotate all possible semantic classes due to the complexity of real-world road scenes.
So we hope CODA can motivate researchers to consider more scalable and effective solutions to build a robust perception system.


\section{Discussion}\label{sec:discussion}

\subsubsection{Effectiveness of COPG.}
\label{sec:effectiveness-COPG}
The examples in Appendix~E qualitatively demonstrates the effectiveness of COPG, reliably identifying nearby objects and retaining corner cases in a progressive manner.
We further quantitatively study the effectiveness of COPG by considering it as a \textit{corner-case detector}, instead of a \textit{corner-case proposal generator}.
The evaluation result is shown in~\cref{tab:ablation}, where COPG is compared with other object and anomaly detectors on detecting the corner cases in CODA-KITTI.
Note that CODA-KITTI is curated by manually examining all the ``misc''-category annotation of KITTI~\cite{Geiger2012CVPR}.
In other words, the construction of CODA-KITTI does not involve COPG.
As reported, COPG shows significant improvements and is much more comparable to human than the baselines.

Moreover, comparing the baseline performances on CODA (\cref{tab:exp_coda}) with those on CODA-ONCE (Tab.~8 in Appendix~D),
we notice all detectors generally achieve higher AR on CODA than CODA-ONCE (\eg, 12.8\% vs 10.2\% AR for BDD100K-trained RetinaNet), suggesting that CODA-ONCE constructed based on the proposals of COPG is much harder than the corner cases of the other two subsets whose construction does not involve COPG. 



\begin{table}[t]
\centering
\caption{Evaluation of COPG and other object/anomaly detectors on detecting corner cases.
The experiments are conducted on CODA-KITTI whose construction does not involve COPG (whereas the construction of CODA-ONCE does).
Here AR$^{m}_{50}$ and AR$^{l}_{50}$ represent AR$_{50}$ for medium and large objects, since no small corner cases are included in CODA-KITTI, with
the same definition for AR$^{m}_{30}$ and AR$^{l}_{30}$ under 0.3 IoU threshold.}
\label{tab:ablation}
\setlength\tabcolsep{5pt}
\begin{tabular}{l|cc|cc}
\toprule
Method & AR$^{m}_{50}$ & AR$^{l}_{50}$ & AR$^{m}_{30}$ & AR$^{l}_{30}$ \\
\hline 
Faster R-CNN~\cite{Ren2015Faster} & 6.7 & 8.3 & 26.4 & 28.8 \\
Memory-based OOD~\cite{memorybankanomaly} & 2.2 & 21.8 & 6.6 & 39.5 \\
Synthesize then Compare~\cite{synthesize} & 9.0 & 17.7 & 12.3 & 33.3 \\
\hline
COPG (Ours) & \textbf{23.8 }& \textbf{44.9} & \textbf{39.6} & \textbf{63.9} \\
\bottomrule
\end{tabular}

\end{table}


\subsubsection{Comparison between closed-world and open-world object detection.}
\begin{figure}[t]
    \centering
    \includegraphics[width=0.9\linewidth]{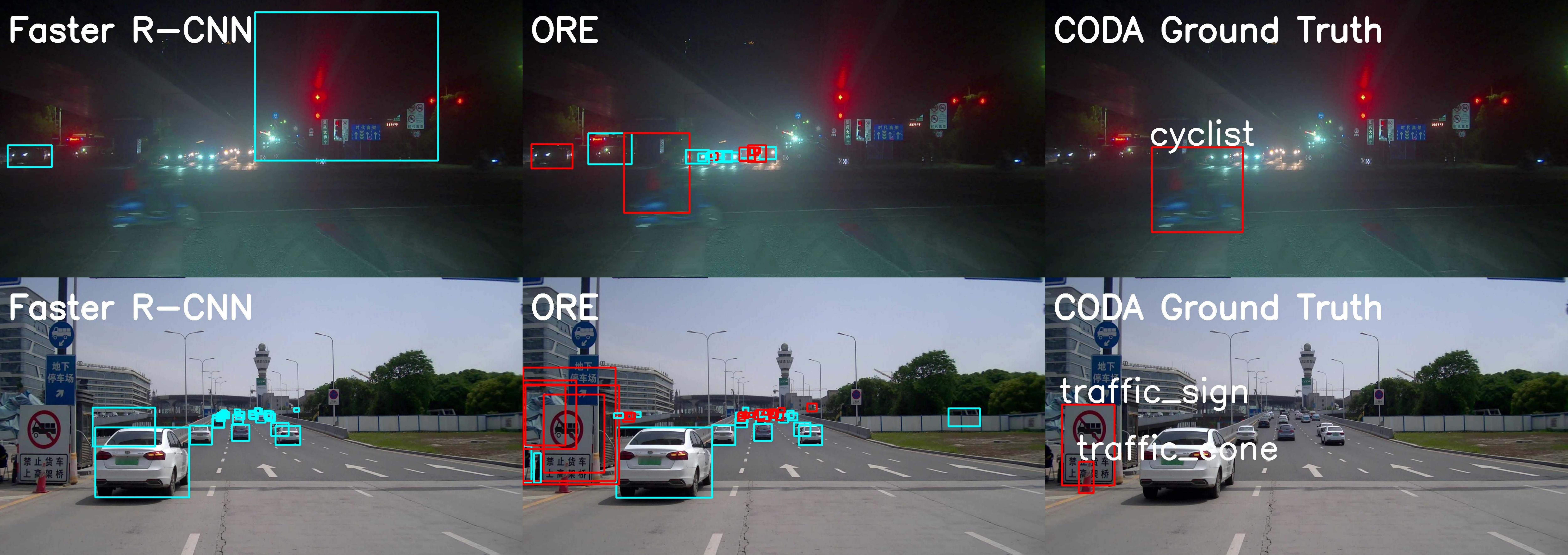}
    \caption{Visualization of Faster R-CNN (left), ORE (middle) detection results and corner case ground truth (right) on CODA.
    We annotate the \emph{unknown} predictions of ORE and CODA ground truth with \emph{red} boxes, while the common-class predictions are annotated by \emph{blue} boxes.
    ORE solves the corner cases of both common (top, cyclist) and novel (bottom, traffic cone \& sign) classes.}
    \label{fig:vis}
\end{figure}

We visualize and compare the detection results of Faster R-CNN, ORE and CODA ground truth in \cref{fig:vis}.
Considering that \emph{unknown} objects are usually trained as background for object detection, ORE utilizes the SODA10M validation set to estimate the known and unknown energy functions based on EBM~\cite{lecun2006tutorial}.
As shown in \cref{fig:vis}, by using an extra data source, ORE can successfully deal with the corner cases of both common and novel classes, which is consistent with the experiment results in \cref{tab:exp_coda} and \cref{tab:exp_once}.
The usage of an extra data source might put ORE at advantage, but the improvement is still impressive since the extra data is only used for the energy function estimation without updating the parameters of the detector at training time.

The performance of ORE does remind us that it is possible to build a more robust perception system by utilizing an additional data source to separate background and \emph{unknown} objects.
However, for ORE, the extra data source is required to be labeled. Considering the annotation cost, it is more desirable to build a system requiring unlabeled data only (\eg, SODA10M large-scale unlabeled set, which has demonstrated to improve cross-domain performance~\cite{han2021soda10m}), of which CODA would be a great help in the evaluation.


\subsubsection{Evaluation of few-shot object detection (FSOD).}
The main goal of CODA is to evaluate the generalization ability of object detectors in self-driving systems \emph{without} model adaptation. Nevertheless, it can also be used to evaluate adaptation methods like FSOD.
So, apart from the typical baselines included in \cref{tab:exp_coda}, we have also evaluated two of the state-of-the-art FSOD methods, FsDet~\cite{wang2020frustratingly} and DeFRCN~\cite{qiao2021defrcn}, on CODA in a 34-way-1-shot setting
with \textbf{5-time} repeated experiments (see \cref{tab:defrcn}).
Neither method demonstrates satisfying performance.


\begin{table}[t]
    \centering
    \setlength\tabcolsep{7.0pt}
    \caption{Evaluation of FSOD on CODA. 
    ``$^{\dag\dag}$'' suggests that the reported values are evaluated in a class-agnostic manner, same as the CORNER setting adopted in \cref{tab:exp_coda}.}
    \label{tab:defrcn}
    \begin{tabular}{c|c|c|c|c|c|c}
        \toprule
        \multirow{2}*{Method} & \multicolumn{3}{c|}{34-way (class-wise)} & \multicolumn{3}{c}{1-way$^{\dag\dag}$ (class-agnostic)} \\
        \cline{2-7}
         & AR & AR$_{50}$ & AR$_{75}$ & AR & AR$_{50}$ & AR$_{75}$ \\
        \hline
        FsDet~\cite{wang2020frustratingly} & $4.9_{\pm 0.8}$ & $9.4 _{\pm 1.9}$ & $4.4_{\pm 0.8}$ & $4.2 _{\pm 0.4}$ & $7.7 _{\pm 0.7}$ & $4.0 _{\pm 0.3}$ \\
        DeFRCN~\cite{qiao2021defrcn} & $6.7 _{\pm 1.2}$ & $12.1 _{\pm 1.6}$ & $6.6 _{\pm 1.6}$ & $4.5 _{\pm 0.5}$ & $8.9 _{\pm 0.9}$ & $4.2 _{\pm 0.5}$ \\
        \bottomrule
    \end{tabular}
\end{table}


\subsubsection{Limitation and potential negative societal impact.} 
See Appendix~C.


\section{Conclusion}

In this paper, we propose CODA, a real-world road corner case dataset for object detection in autonomous driving, constructed by ground truth class separation and automatic proposal.
We observe a significant performance drop for state-of-the-art detectors when deployed on CODA. 
We further provide a thorough comparison of different methods and shed light on potential solutions to a more robust perception system.
We hope that CODA can motivate further research in reliable detection for real-world autonomous driving.

\section*{Acknowledgement}
We gratefully acknowledge the support of MindSpore, CANN (Compute Architecture for Neural Networks) and Ascend AI Processor used for this research.



\clearpage
%
%
\bibliographystyle{splncs04}
\bibliography{egbib}

\clearpage
\appendix
\section*{Appendix}
The official website of CODA is at \url{https://coda-dataset.github.io}, where we have released 1000 of the 1500 annotated scenes of CODA.
The remaining 500 scenes are reserved for the corner case challenge in the SSLAD workshop\footnote{\url{https://ssladcompetition.github.io/}} of ECCV 2022, and will be released after the challenge.

As a continued effort, CODA is further extended by 8711 additional scenes with more than 28k new corner cases during the reviewing and publishing process of this paper. The extension will also be made fully available after the ECCV challenge. Please stay tuned!


\section{Supplementary implementation details}\label{app:implement}

\subsection{Closed-world detectors}
We re-implement the closed-world detectors in Tab.~\ref{tab:exp_coda} following the default configurations in MMDetection~\cite{mmdetection} if the corresponding official checkpoints\footnote{SODA10M: \url{https://soda-2d.github.io/} \& BDD100K: \url{https://github.com/SysCV/bdd100k-models/}} are not publicly available.
Detectors are by default trained for 12 epochs (1x) on BDD100K and Waymo, while for 24 epochs (2x) on SODA10M, except Deformable DETR and Sparse R-CNN, which are trained for 100 epochs on SODA10M training set, due to the limited labeled data size of SODA10M and data-hungry property of the Transformer layers.


\subsection{ORE}
The experiments are conducted using the released code\footnote{\url{https://github.com/JosephKJ/OWOD}} of the original paper~\cite{joseph2021towards}, without incremental learning.
We use the training set of SODA10M~\cite{han2021soda10m} for training; the validation set of SODA10M and the whole CODA for fitting the energy distributions; the test set of SODA10M and the whole CODA for testing.
The backbone of the network is ResNet-50~\cite{he2016deep}, and the whole network is trained for 24 epochs with a batch size of 8 under a learning rate of 0.02.
More efficient training methods (\eg, sparse training~\cite{zhou2021effective}) are potential solutions for better generalization, which will be explored in the future.


\subsection{Anomaly detection}
We utilize the segmentation model and GAN pre-trained on Cityscapes dataset~\cite{cityscapes_dataset} in method \emph{synthesize then compare} during the test procedure. We crop the generated fake images and the original images according to the top-ranking bounding box proposals obtained from RPN, then we compare the cropped images by pixel-wise cosine similarity as illustrated in the paper and get the detection result according to the similarity rankings.
As to \emph{memory-based OOD detection}, we utilize the feature map of common ground truth and some conventional background (vegetation, sky, and pure black) extracted from ResNet-152~\cite{he2016deep} as the memory bank. Likewise, We then compare the feature map extracted from off-the-shelf ResNet-152 of the top-ranking bounding boxes from RPN with the memory bank to filter out the anomaly detection results.
More advanced OOD detection methods~\cite{zhou22model,zhou22sparse} will be explored in the future.


\section{Supplementary ablation studies on COPG}
\label{app:ablation}
We conduct ablation studies on COPG by tuning its components one at a time.
The experiment is conducted on CODA-ONCE whose final ground truths are utilized to compute the AP and AR (under COCO~\cite{coco_dataset} protocol) of the output proposals of the modified COPGs.
The results are shown in \cref{tab:ablation-cluster}-\ref{tab:ablation-det-iou}, where the \textbf{bold} hyperparameters are the ones in effect.
The number of proposals (\#Proposals) and the number of scenes (\#Scenes) containing at least one proposal are also reported.


\begin{table}[t]
    \centering
    \caption{Ablation study on point-cloud clustering. The min$.$ $\theta$ controls the minimum tolerance of the angle between two points being assigned to the same object (see \cref{fig:clustering}). The cluster size controls the minimum tolerance to the number of points of a cluster being considered as a non-trivial object. The max$.$ dist$.$ controls the maximum tolerance to the distance from the nearest point of a cluster to the lidar sensor.}
    \label{tab:ablation-cluster}
    \begin{tabular}{ccccc}
        \toprule
        \hphantom{\quad}Min. $\theta$\hphantom{\quad} & AP & AR & \#Proposals & \#Scenes \\
        \midrule
        4$^{\circ}$  & 1.3 & 3.1 & 1039 & 763  \\
        \textbf{8$^{\circ}$} & 2.0 & 5.1 & 1522 & 1040 \\
        12$^{\circ}$ & 1.9 & 5.0 & 1620 & 1000 \\
        16$^{\circ}$ & 1.7 & 4.7 & 1652 & 965  \\
        \bottomrule
    \end{tabular}
    \vskip 2mm
    \begin{tabular}{ccccc}
        \toprule
        Cluster size & AP & AR & \#Proposals & \#Scenes \\
        \midrule
        5           & 2.0 & 5.1 & 1522 & 1040 \\
        \textbf{10} & 2.0 & 5.1 & 1522 & 1040 \\
        20          & 2.0 & 5.1 & 1513 & 1038 \\
        \bottomrule
    \end{tabular}
    \vskip 2mm
    \begin{tabular}{ccccc}
        \toprule
        \hphantom{ }Max. dist.\hphantom{ }\, & AP & AR & \#Proposals & \#Scenes \\
        \midrule
        25          & 1.8 & 4.5 & 1336 & 935  \\
        \textbf{50} & 2.0 & 5.1 & 1522 & 1040 \\
        100         & 1.9 & 5.1 & 1548 & 1040 \\
        \bottomrule
    \end{tabular}
    \vskip 4mm
    \caption{Ablation study on background removal. The background (BG) ratio controls the maximum tolerance to the ratio of the proposal area overlapped with the background regions.}
    \label{tab:ablation-seg-iou}
    \begin{tabular}{ccccc}
        \toprule
        BG ratio & AP & AR & \#Proposals & \#Scenes \\
        \midrule
        0.15          & 0.4 & 1.1 & 412  & 308  \\
        0.30          & 1.0 & 2.6 & 834  & 614  \\
        \textbf{0.45} & 2.0 & 5.1 & 1522 & 1040 \\
        0.60          & 2.0 & 5.6 & 1834 & 1055 \\
        0.75          & 1.8 & 6.1 & 2315 & 1056 \\
        \bottomrule
    \end{tabular}
    \vskip 4mm
    \caption{Ablation study on common-class suppression. The IoU threshold controls the maximum tolerance to the IoU between every proposal and each detected common-class object.}
    \label{tab:ablation-det-iou}
    \begin{tabular}{ccccc}
        \toprule
        IoU & AP & AR & \#Proposals & \#Scenes \\
        \midrule
        0.00          & 1.2 & 2.7 & 834  & 643  \\
        \textbf{0.25} & 2.0 & 5.1 & 1522 & 1040 \\
        0.50          & 1.8 & 5.2 & 1660 & 1041 \\
        \bottomrule
    \end{tabular}
\end{table}






\section{Limitation and potential negative societal impact}\label{app:discuss}
We would continue to enlarge CODA by exploring:
1) Use COPG on more real-world road scenes.
2) Since CODA is collected in the real world with high-quality annotation, we can generate more synthesized images following~\cite{blum2019fishyscapes,hendrycks2019benchmark}, or mine large-scale unlabeled road scene images in a semi-supervised manner~\cite{qiao2021defrcn,reza2019automatic,sohn2020detection,liu2021unbiased}.

CODA has no potential negative societal impact since it is constructed totally based on publicly available datasets with delicate privacy protection.
For example, all objects containing personal information (\eg, human faces, license plates) are blurred in ONCE~\cite{mao2021one}.


\clearpage
\section{Supplementary benchmark results}\label{app:exp-baseline}


\begin{table}[h]
    \centering
    \caption{Detection results (\%) on our CODA-ONCE dataset.
    The dramatic performance decrease still maintains, but compared with \cref{tab:exp_coda}, we observe a decrease for the reported AR values of all detectors, suggesting that CODA-ONCE constructed with our automatic corner case proposal generation COPG in \cref{sec:COPG} is the most challenging subset of CODA.
    }
    \label{tab:exp_once}
    \setlength\tabcolsep{2.2pt}
    \scalebox{0.75}{
    \begin{tabular}{l|c|cccc|cccc|cccc}
    \toprule 
    \multicolumn{2}{c|}{CODA-ONCE} & \multicolumn{4}{c|}{CORNER} & \multicolumn{4}{c|}{COMMON} & \multicolumn{4}{c}{NOVEL} \\
    \hline
    Method & Dataset & AR$^*$ & AR$_{50}$ & AR$_{75}$ & AR$^{10}$ & AR$^*$ & AR$_{50}$ & AR$_{75}$ & AR$^{10}$ & AR$^*$ & AR$_{50}$ & AR$_{75}$ & AR$^{10}$ \\
    \hline 
    RetinaNet~\cite{Lin2017Focal} & & \textbf{9.6} & 21.8 & 7.0 & 3.0 & 29.8 & 59.8 & 24.5 & 24.8 & - & - & - & -  \\
    Faster R-CNN~\cite{Ren2015Faster} & & 4.0 & 8.5 & 3.4 & 2.3 & 23.3 & 44.3 & 19.6 & 22.5 & - & - & - & - \\
    Cascade R-CNN~\cite{cai18cascadercnn} & & 5.6 & 11.4 & 4.9 & 2.9 & 27.2  & 46.0  & 30.7 & 25.3 & - & - & - & - \\
    Deformable DETR~\cite{zhu2020deformable} & & 6.1 & 13.0 & 5.3 & 2.7 & \textbf{36.8} & 63.9 & 40.0 & 31.2 & - & - & - & -  \\
    Sparse R-CNN~\cite{sun2021sparse} & SODA10M & 4.4 & 9.2 & 3.7 & 2.1 & 26.3 & 44.5 & 25.5 & 22.7 & - & - & - & -  \\
    Cascade Swin~\cite{liu2021swin} & \cite{han2021soda10m} & 5.5 & 11.3 & 4.8 & 2.9 & 31.2 & 49.1 & 35.5 & 30.0  & - & - & - & -  \\
    RPN (Faster)~\cite{Ren2015Faster} & & 5.1 & 11.3  & 4.0 & 1.3 & - & - & - & - & - & - & - & - \\
    RPN (Cascade)~\cite{cai18cascadercnn} & & 4.8 & 11.4  & 3.5 & 1.3  & - & - & - & - & - & - & - & - \\
    ORE~\cite{joseph2021towards} & & 5.3 & 11.5 & 4.4 & 2.8 & 19.2 & 36.8  & 17.1 & 18.6 & \textbf{2.0} & 4.8	& 1.6 & 1.7 \\
    \hline 
    RetinaNet~\cite{Lin2017Focal} & & \textbf{10.2} & 19.4 & 9.1 & 2.6 & 30.1 & 63.7 & 23.3 & 25.8 & \textbf{7.3} & 14.1 & 6.6 & 3.8 \\
    Faster R-CNN~\cite{Ren2015Faster} & & 7.8 & 15.1  & 7.4 & 2.3 & 26.8  & 51.8  & 23.2  & 24.3 & 5.5   & 10.7  & 5.2 & 4.1 \\
    Cascade R-CNN~\cite{cai18cascadercnn} & & 7.7 & 14.6  & 6.9 & 2.4 & 26.7  & 48.8  & 24.0  & 24.6 & 5.7   & 10.5  & 5.1 & 4.3 \\
    Deformable DETR~\cite{zhu2020deformable} & BDD100K & 7.8 & 18.4 & 5.6 & 1.9 & 30.9 & 67.8 & 24.4 & 28.5 & 5.8 & 13.4 & 4.2 & 3.0  \\
    Sparse R-CNN~\cite{sun2021sparse} & \cite{bdd100k} & 7.6 & 15.4 & 6.7 & 2.6 & 30.3 & 56.0 & 28.8 & 26.9 & 5.9 & 12.0 & 5.3 & 3.4  \\
    Cascade Swin~\cite{liu2021swin} &  & 7.2 & 13.3 & 6.8 & 2.9 & \textbf{32.9} & 57.0 & 33.0 & 31.3 & 5.3 & 9.9 & 5.0 & 4.5  \\
    RPN (Faster)~\cite{Ren2015Faster} &  & 7.9   & 16.1  & 7.0 & 1.6  & - & - & - & -  & - & - & - & - \\
    RPN (Cascade)~\cite{cai18cascadercnn} & & 7.7 & 15.8  & 6.9 & 1.8  & - & - & - & -  & - & - & - & - \\
    \hline 
    RetinaNet~\cite{Lin2017Focal} & & 5.4 & 11.1 & 4.5 & 2.2 & 26.1 & 45.0 & 27.0 & 23.9 & 3.0 & 6.1 & 2.7 & 1.4 \\
    Faster R-CNN~\cite{Ren2015Faster} & & 4.1 & 8.3 & 3.5 & 2.1 & 22.9 & 38.9 & 21.9 & 20.8 & 2.3 & 4.9 & 2.0 & 1.4 \\
    Cascade R-CNN~\cite{cai18cascadercnn} & & 3.8 & 7.1 & 3.6 & 2.2 & 20.0 & 34.1 & 21.8 & 18.7 & 2.2 & 3.8 & 2.2 & 1.5 \\
    Deformable DETR~\cite{zhu2020deformable} & Waymo & 5.5 & 11.4 & 4.6 & 1.9 & 30.8 & 52.9 & 26.8 & 24.0 & 3.1 & 6.5 & 2.8 & 1.1  \\
    Sparse R-CNN~\cite{sun2021sparse} & \cite{Sun2020Scalability} & \textbf{7.3} & 14.6 & 6.4 & 2.1 & \textbf{32.4} & 55.4 & 30.0 & 23.9 & \textbf{4.2} & 8.6 & 3.8 & 1.4  \\
    Cascade Swin~\cite{liu2021swin} &  & 2.5 & 4.4 & 2.4 & 1.8 & 23.5 & 41.1 & 20.7 & 23.0 & 1.2 & 2.0 & 1.2 & 0.8  \\
    RPN (Faster)~\cite{Ren2015Faster} &  & 4.6 & 9.4 & 4.3 & 1.5  & - & - & - & -  & - & - & - & - \\
    RPN (Cascade)~\cite{cai18cascadercnn} & & 4.3 & 9.2 & 3.6 & 1.6  & - & - & - & -  & - & - & - & - \\
    \bottomrule 
    \end{tabular}}
\end{table}

\begin{table}
    \centering
    \caption{More detection results(\%) on our CODA-ONCE dataset.
    AR$^s$, AR$^m$ and AR$^l$ are the average recall for small, medium and large objects respectively, following COCO definition~\cite{coco_dataset}, while AR$^1$ represents the average recall when only 1 object prediction is allowed for each image.
    Here $AR^s$ is also marked as ``-'', since no small corner cases of common classes are collected in CODA-ONCE subset.}
    \label{tab:app_once}
    \setlength\tabcolsep{2.2pt}
    \scalebox{0.8}{
    \begin{tabular}{l|c|cccc|cccc|cccc}
    \toprule 
    \multicolumn{2}{c|}{CODA-ONCE} & \multicolumn{4}{c|}{CORNER} & \multicolumn{4}{c|}{COMMON} & \multicolumn{4}{c}{NOVEL} \\
    \hline
    Method & Dataset & AR$^{s}$ & AR$^{m}$ & AR$^{l}$ & AR$^{1}$ &  AR$^{s}$ & AR$^{m}$ & AR$^{l}$ & AR$^{1}$ & AR$^{s}$ & AR$^{m}$ & AR$^{l}$ & AR$^{1}$ \\
    \hline 
    RetinaNet~\cite{Lin2017Focal} & & 3.8 & 2.9 & 19.4  & 0.2 & - & 0.0   & 33.4 & 4.7 & - & - & - & - \\ 
    Faster R-CNN~\cite{Ren2015Faster} & & 0.6   & 1.0   & 8.5 & 0.2 & - & 0.0   & 26.0 & 3.7 & - & - & - & -  \\
    Cascade R-CNN~\cite{cai18cascadercnn} & & 1.0   & 1.5   & 11.9 & 0.2 & - & 0.0   & 30.8 & 4.9 & - & - & - & -  \\
    Deformable DETR~\cite{zhu2020deformable} & & 0.7 & 1.9 & 12.6 & 0.2 & - & 1.1 & 42.3 & 6.1 & - & - & - & -  \\
    Sparse R-CNN~\cite{sun2021sparse} & SODA10M & 1.3 & 1.6 & 8.6 & 0.3 & - & 3.3 & 28.6 & 7.2 & - & - & - & -  \\
    Cascade Swin~\cite{liu2021swin} & \cite{han2021soda} & 1.1 & 1.5 & 11.5 & 0.2 & - & 0.0 & 35.8 & 3.9  & - & - & - & -  \\
    RPN (Faster)~\cite{Ren2015Faster} & & 2.2  & 1.5   & 10.3 & 0.3 & - & - & - & - & - & - & - & -  \\
    RPN (Cascade)~\cite{cai18cascadercnn} &  & 3.6   & 1.5   & 9.3 & 0.3 & - & - & - & - & - & - & - & -  \\
    ORE~\cite{joseph2021towards} & & 1.6 & 1.9 & 10.4 & 0.3 & - & 1.7 & 21.1 & 5.3 & 0.5 & 0.6 & 4.4 & 0.4 \\
    \hline 
    RetinaNet~\cite{Lin2017Focal} & & 2.2   & 4.4   & 19.2  & 0.2 & -   & 9.2   & 33.1  & 5.9 & 1.1   & 3.4   & 13.9  & 0.6 \\
    Faster R-CNN~\cite{Ren2015Faster} & & 1.3   & 2.6   & 15.9  & 0.1 & -   & 17.2  & 29.3  & 4.7 & 0.4   & 2.2   & 11.2  & 0.5 \\
    Cascade R-CNN~\cite{cai18cascadercnn} & & 1.7   & 2.6   & 15.5  & 0.1 & -   & 12.5  & 30.4  & 5.4 & 0.8   & 2.2   & 11.5  & 0.6 \\
    Deformable DETR~\cite{zhu2020deformable} & BDD100K & 1.3 & 3.6 & 14.5 & 0.1 & - & 18.9 & 33.1 & 7.2 & 1.0 & 3.0 & 10.8 & 0.4  \\
    Sparse R-CNN~\cite{sun2021sparse} & \cite{bdd100k} & 1.7 & 3.3 & 14.2 & 0.2 & - & 4.4 & 34.3 & 5.4 & 1.3 & 2.9 & 11.0 & 0.5  \\
    Cascade Swin~\cite{liu2021swin} &  & 1.3 & 2.4 & 14.5 & 0.3 & - & 27.5 & 37.3 & 9.6 & 0.7 & 2.0 & 10.8 & 0.8  \\
    RPN (Faster)~\cite{Ren2015Faster} &  & 2.2   & 3.0   & 15.3  & 0.2 & - & - & - & - & - & - & - & - \\
    RPN (Cascade)~\cite{cai18cascadercnn} & & 1.9   & 2.9   & 14.9  & 0.2  & - & - & - & - & - & - & - & - \\
    \hline 
    RetinaNet~\cite{Lin2017Focal} & & 5.4 & 11.1 & 4.5 & 2.2 & 26.1 & 45.0 & 27.0 & 23.9 & 3.0 & 6.1 & 2.7 & 1.4 \\
    Faster R-CNN~\cite{Ren2015Faster} & & 4.1 & 8.3 & 3.5 & 2.1 & 22.9 & 38.9 & 21.9 & 20.8 & 2.3 & 4.9 & 2.0 & 1.4 \\
    Cascade R-CNN~\cite{cai18cascadercnn} & & 3.8 & 7.1 & 3.6 & 2.2 & 20.0 & 34.1 & 21.8 & 18.7 & 2.2 & 3.8 & 2.2 & 1.5 \\
    Deformable DETR~\cite{zhu2020deformable} & Waymo & 5.5 & 11.4 & 4.6 & 1.9 & 30.8 & 52.9 & 26.8 & 24.0 & 3.1 & 6.5 & 2.8 & 1.1  \\
    Sparse R-CNN~\cite{sun2021sparse} & \cite{Sun2020Scalability} & \textbf{7.3} & 14.6 & 6.4 & 2.1 & \textbf{32.4} & 55.4 & 30.0 & 23.9 & \textbf{4.2} & 8.6 & 3.8 & 1.4  \\
    Cascade Swin~\cite{liu2021swin} &  & 2.5 & 4.4 & 2.4 & 1.8 & 23.5 & 41.1 & 20.7 & 23.0 & 1.2 & 2.0 & 1.2 & 0.8  \\
    RPN (Faster)~\cite{Ren2015Faster} &  & 4.6 & 9.4 & 4.3 & 1.5  & - & - & - & -  & - & - & - & - \\
    RPN (Cascade)~\cite{cai18cascadercnn} & & 4.3 & 9.2 & 3.6 & 1.6  & - & - & - & -  & - & - & - & - \\
    \bottomrule 
    \end{tabular}}
\end{table}

\begin{table}
    \centering
    \caption{More detection results(\%) on our CODA dataset.
    AR$^s$, AR$^m$ and AR$^l$ are the average recall for small, medium and large objects respectively, following COCO definition~\cite{coco_dataset}, while AR$^1$ represents the average recall when only 1 object prediction is allowed for each image.}
    \label{tab:app_coda}
    \setlength\tabcolsep{2.2pt}
    \scalebox{0.8}{
    \begin{tabular}{l|c|cccc|cccc|cccc}
    \toprule 
    \multicolumn{2}{c|}{CODA} & \multicolumn{4}{c|}{CORNER} & \multicolumn{4}{c|}{COMMON} & \multicolumn{4}{c}{NOVEL} \\
    \hline
    Method & Dataset & AR$^{s}$ & AR$^{m}$ & AR$^{l}$ & AR$^{1}$ & AR$^{s}$ & AR$^{m}$ & AR$^{l}$ & AR$^{1}$ & AR$^{s}$ & AR$^{m}$ & AR$^{l}$ & AR$^{1}$ \\
    \hline 
    RetinaNet~\cite{Lin2017Focal} & & 5.6   & 6.5   & 20.6 & 0.2  & 40.0  & 1.3   & 31.8 & 5.0 & - & - & - & -  \\ 
    Faster R-CNN~\cite{Ren2015Faster} & & 2.7   & 4.6   & 10.8 & 0.2 & 60.0  & 4.7   & 24.8 & 5.6 & - & - & - & - \\
    Cascade R-CNN~\cite{cai18cascadercnn} & & 3.3   & 5.1   & 13.7  & 0.2 & 60.0  & 4.7   & 28.6 & 6.2 & - & - & - & - \\
    Deformable DETR~\cite{zhu2020deformable} & & 1.9 & 3.5 & 13.6 & 0.2 & 0.0 & 6.7 & 38.8 & 6.5 & - & - & - & -  \\
    Sparse R-CNN~\cite{sun2021sparse} & SODA10M & 1.7 & 4.4 & 10.2 & 0.5 & 70.0 & 5.3 & 27.2 & 8.3 & - & - & - & -  \\
    Cascade Swin~\cite{liu2021swin} & \cite{han2021soda10m} & 3.3 & 5.3 & 13.2 & 0.3 & 50.0 & 4.0 & 33.7 & 4.7  & - & - & - & -  \\
    RPN (Faster)~\cite{Ren2015Faster} & & 5.1   & 5.7   & 11.9  & 0.6 & - & - & - & - & - & - & - & - \\
    RPN (Cascade)~\cite{cai18cascadercnn} &  & 4.8   & 5.3   & 11.6 & 0.5 & - & - & - & - & - & - & - & -  \\
    ORE~\cite{joseph2021towards} & & 4.0 & 5.5 & 13.0 & 0.3 & 0.0 & 1.3 & 20.2 & 5.5 & 1.4 & 1.9 & 5.9 & 2.9 \\
    \hline 
    RetinaNet~\cite{Lin2017Focal} & & 4.1   & 8.3   & 21.0  & 0.1  & 0.0   & 7.7   & 30.1  & 5.4  & 3.1   & 6.5   & 15.9   & 1.2  \\
    Faster R-CNN~\cite{Ren2015Faster} & & 3.4   & 6.6   & 17.8  & 0.1 & 10.0  & 15.7  & 26.3  & 4.3 & 2.2   & 4.4   & 12.3   & 0.9  \\
    Cascade R-CNN~\cite{cai18cascadercnn} & & 3.3   & 6.6   & 17.2  & 0.1 & 60.0  & 15.0  & 27.5  & 5.8  & 1.7   & 4.1   & 12.3  & 1.1  \\
    Deformable DETR~\cite{zhu2020deformable} & BDD100K & 2.1 & 5.4 & 15.5 & 0.1 & 0.0 & 17.7 & 29.9 & 7.1 & 1.7 & 4.4 & 12.0 & 0.5  \\
    Sparse R-CNN~\cite{sun2021sparse} & \cite{bdd100k} & 3.2 & 6.6 & 15.6 & 0.2 & 0.0 & 4.0 & 30.7 & 4.9 & 2.8 & 5.7 & 12.7 & 1.1  \\
    Cascade Swin~\cite{liu2021swin} &  & 3.4 & 6.3 & 16.4 & 0.2 & 0.0 & 27.0 & 35.5 & 9.8 & 2.0 & 3.9 & 11.4 & 1.6  \\
    RPN (Faster)~\cite{Ren2015Faster} & & 4.3   & 6.9   & 17.2  & 0.5 & - & - & - & - & - & - & - & - \\
    RPN (Cascade)~\cite{cai18cascadercnn} & & 3.8   & 6.9   & 17.1  & 0.4 & - & - & - & - & - & - & - & - \\
    \hline 
    RetinaNet~\cite{Lin2017Focal} & & 1.2 & 5.6 & 14.0 & 0.2 & 30.0 & 4.0 & 27.1 & 4.5 & 0.6 & 4.8 & 10.9 & 0.0 \\
    Faster R-CNN~\cite{Ren2015Faster} & & 1.1 & 4.5 & 11.3 & 0.2 & 0.0 & 50.0 & 23.5 & 4.2 & 0.7 & 4.1 & 8.6 & 0.1 \\
    Cascade R-CNN~\cite{cai18cascadercnn} & & 1.2 & 4.5 & 10.8 & 0.1 & 0.0 & 10.0 & 20.5 & 3.7 & 0.6 & 4.2 & 8.2 & 0.1 \\
    Deformable DETR~\cite{zhu2020deformable} & Waymo & 2.1 & 4.9 & 11.7 & 0.2 & 0.0 & 8.0 & 30.8 & 2.6 & 1.2 & 3.8 & 8.3 & 0.1  \\
    Sparse R-CNN~\cite{sun2021sparse} & \cite{Sun2020Scalability} & 3.3 & 7.5 & 15.2 & 0.4 & 0.0 & 4.0 & 33.2 & 4.7 & 2.4 & 5.8 & 11.6 & 0.3  \\
    Cascade Swin~\cite{liu2021swin} &  & 0.3 & 3.7 & 8.8 & 0.3 & 0.0 & 8.3 & 24.5 & 2.8 & 0.2 & 3.5 & 6.4 & 0.2  \\
    RPN (Faster)~\cite{Ren2015Faster} &  & 1.3 & 5.1 & 12.3 & 0.5  & - & - & - & -  & - & - & - & - \\
    RPN (Cascade)~\cite{cai18cascadercnn} & & 1.0 & 5.0 & 12.3 & 0.4  & - & - & - & -  & - & - & - & - \\
    \bottomrule 
    \end{tabular}}
\end{table}


\clearpage
\section{COPG examples}\label{app:copg-examples}
\begin{figure}[H]
    \centering
    \begin{minipage}[t]{\linewidth}
        \centering
        \includegraphics[width=0.1\linewidth]{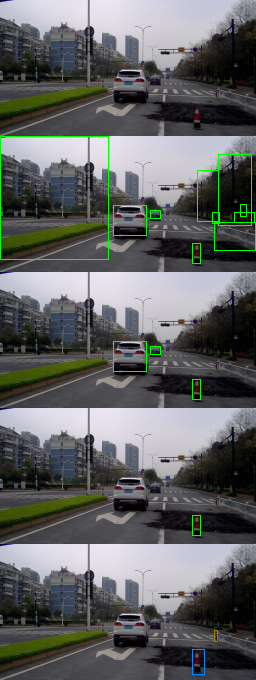}
        \includegraphics[width=0.1\linewidth]{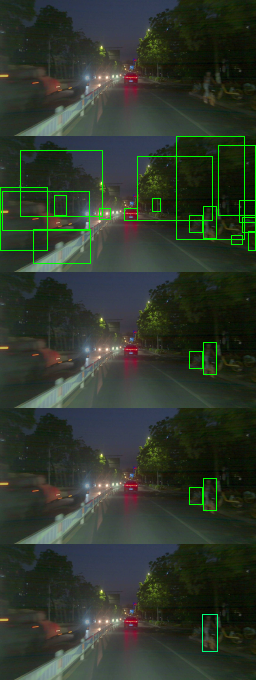}
        \includegraphics[width=0.1\linewidth]{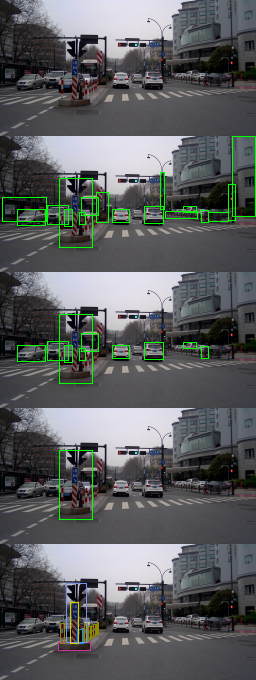}
        \includegraphics[width=0.1\linewidth]{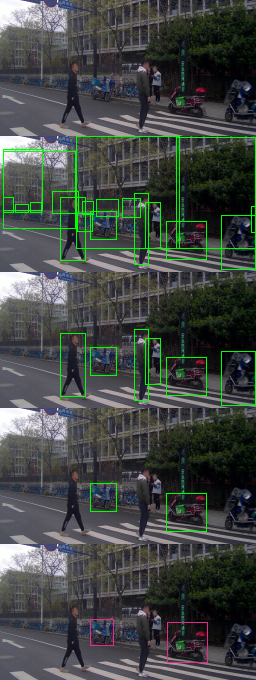}
        \includegraphics[width=0.1\linewidth]{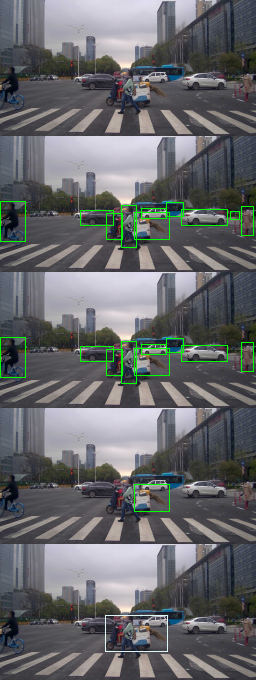}
        \includegraphics[width=0.1\linewidth]{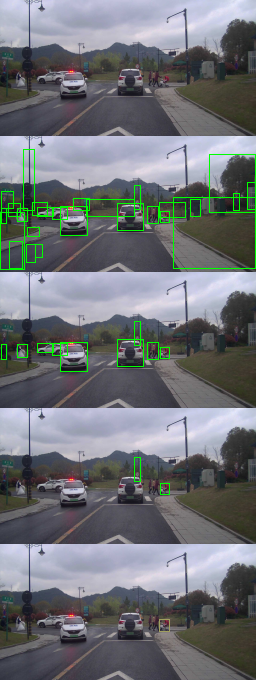}
        \includegraphics[width=0.1\linewidth]{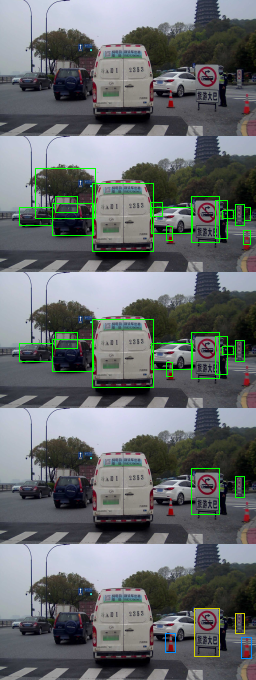}
        \includegraphics[width=0.1\linewidth]{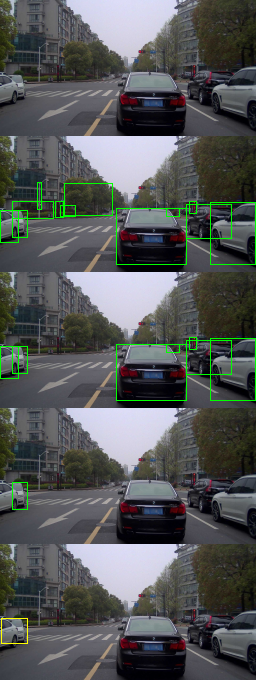}
        \includegraphics[width=0.1\linewidth]{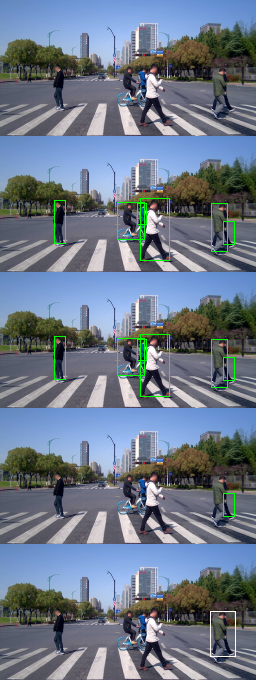}
    \end{minipage}
    \begin{minipage}[t]{\linewidth}
        \centering
        \includegraphics[width=0.1\linewidth]{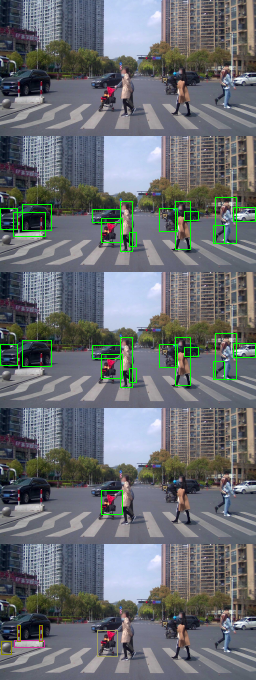}
        \includegraphics[width=0.1\linewidth]{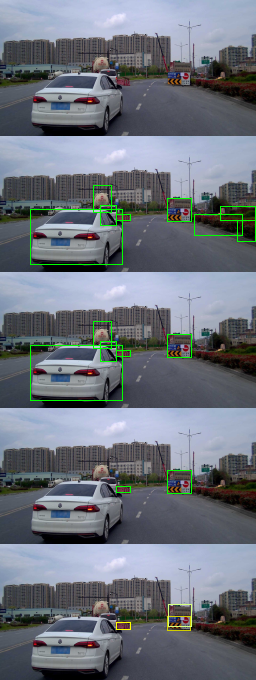}
        \includegraphics[width=0.1\linewidth]{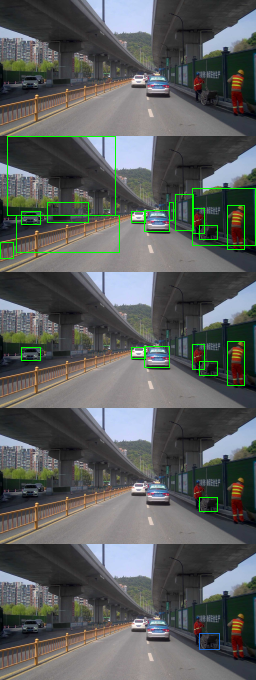}
        \includegraphics[width=0.1\linewidth]{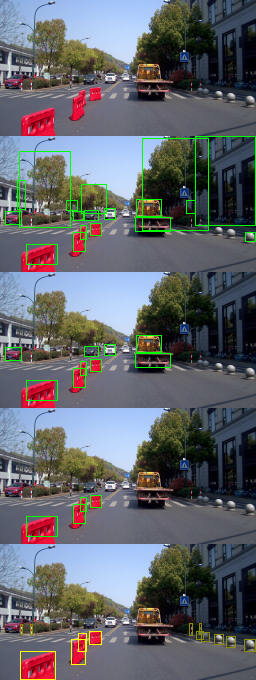}
        \includegraphics[width=0.1\linewidth]{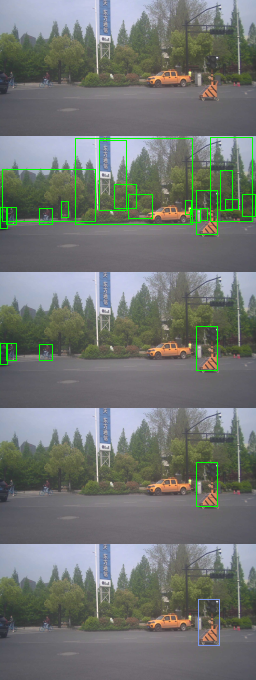}
        \includegraphics[width=0.1\linewidth]{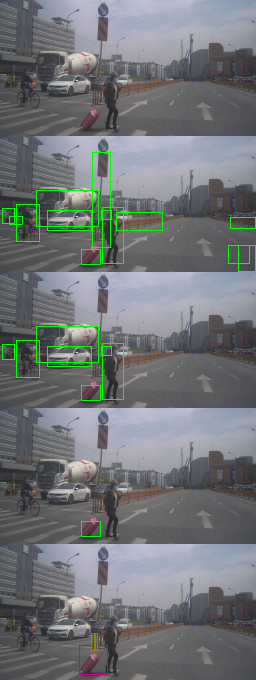}
        \includegraphics[width=0.1\linewidth]{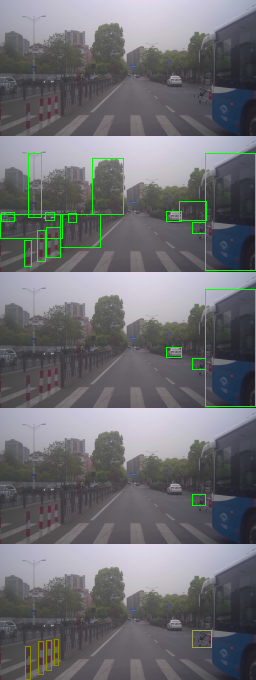}
        \includegraphics[width=0.1\linewidth]{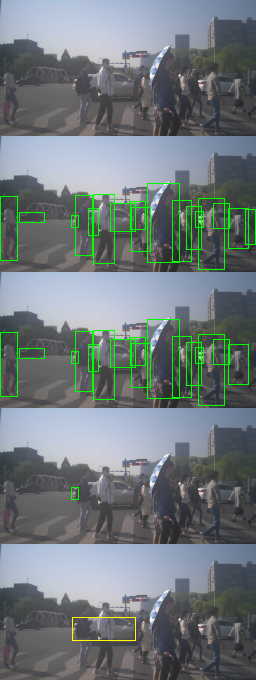}
        \includegraphics[width=0.1\linewidth]{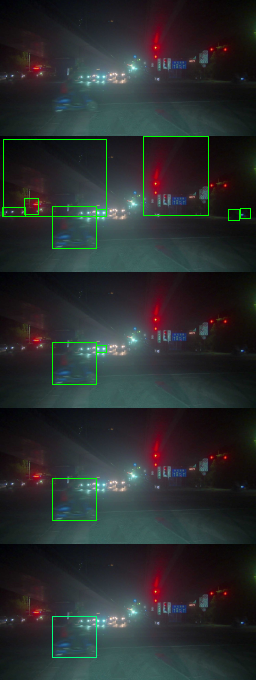}
    \end{minipage}
    \begin{minipage}[t]{\linewidth}
        \centering
        \includegraphics[width=0.1\linewidth]{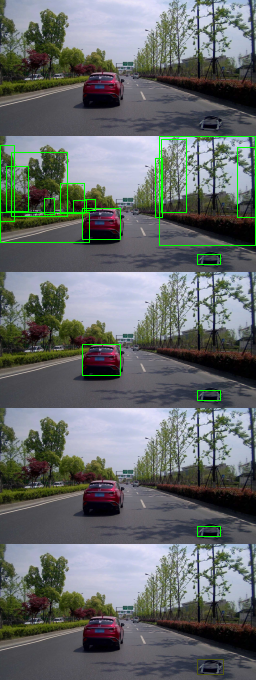}
        \includegraphics[width=0.1\linewidth]{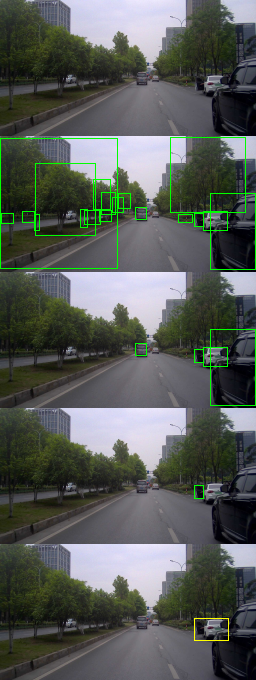}
        \includegraphics[width=0.1\linewidth]{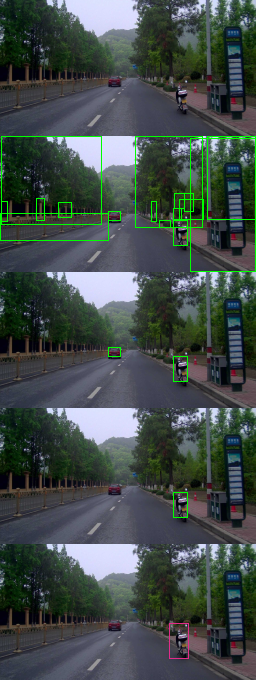}
        \includegraphics[width=0.1\linewidth]{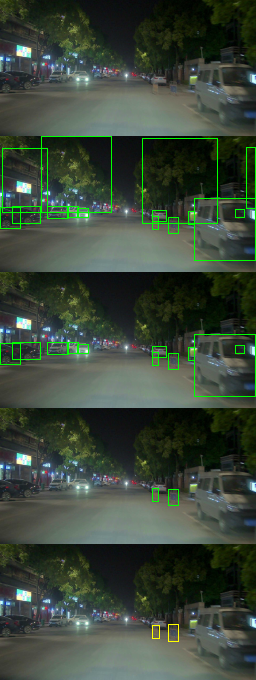}
        \includegraphics[width=0.1\linewidth]{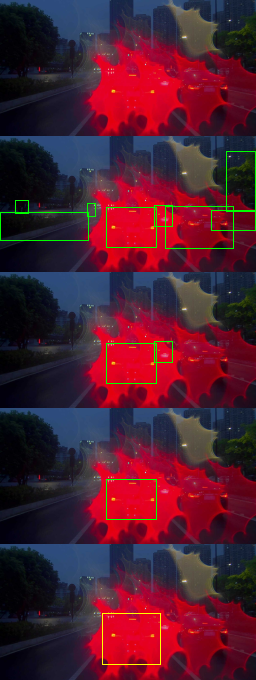}
        \includegraphics[width=0.1\linewidth]{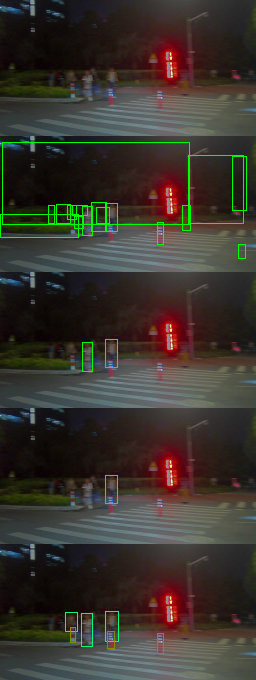}
        \includegraphics[width=0.1\linewidth]{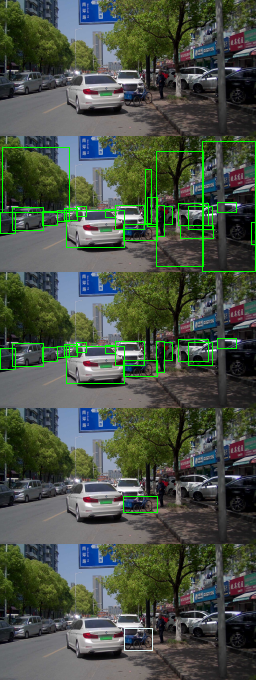}
        \includegraphics[width=0.1\linewidth]{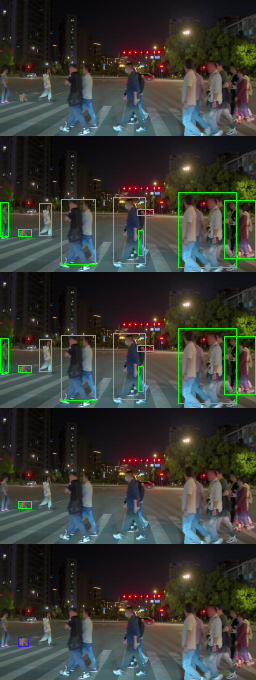}
        \includegraphics[width=0.1\linewidth]{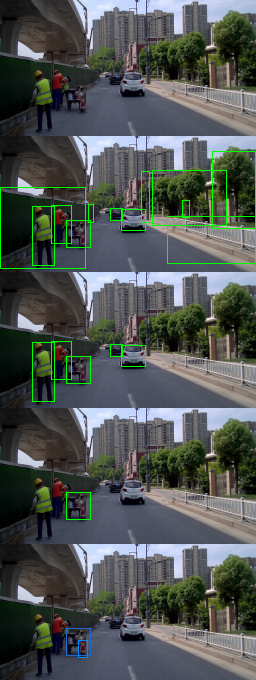}
    \end{minipage}
    \begin{minipage}[t]{\linewidth}
        \centering
        \includegraphics[width=0.1\linewidth]{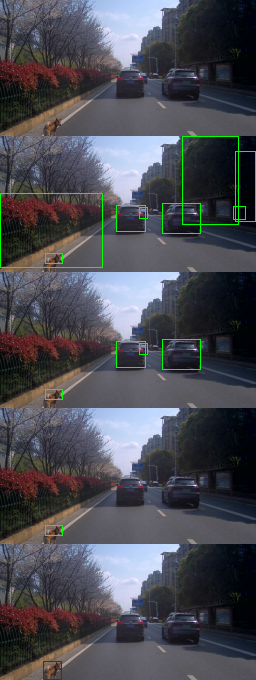}
        \includegraphics[width=0.1\linewidth]{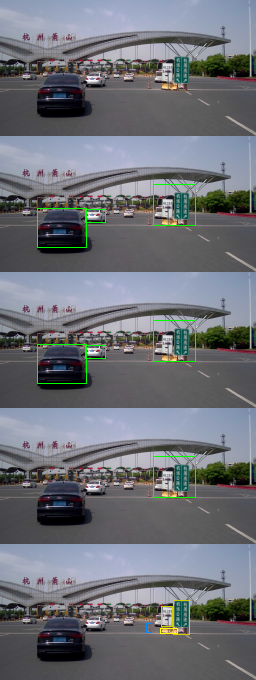}
        \includegraphics[width=0.1\linewidth]{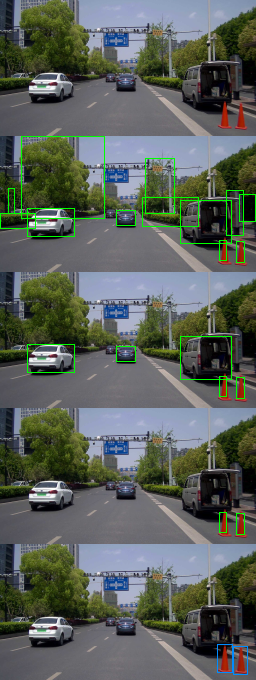}
        \includegraphics[width=0.1\linewidth]{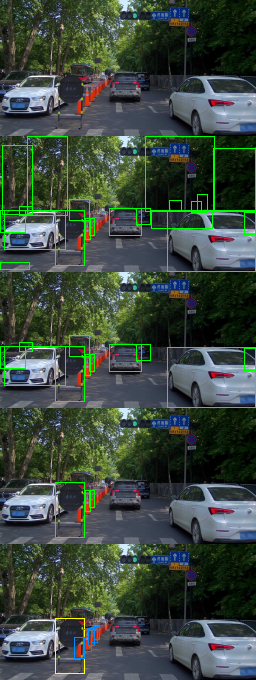}
        \includegraphics[width=0.1\linewidth]{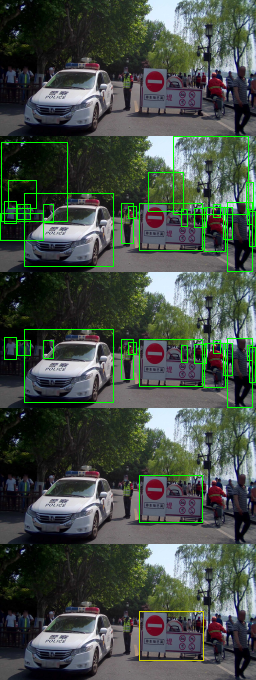}
        \includegraphics[width=0.1\linewidth]{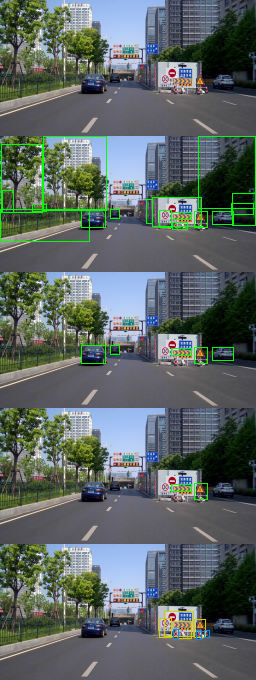}
        \includegraphics[width=0.1\linewidth]{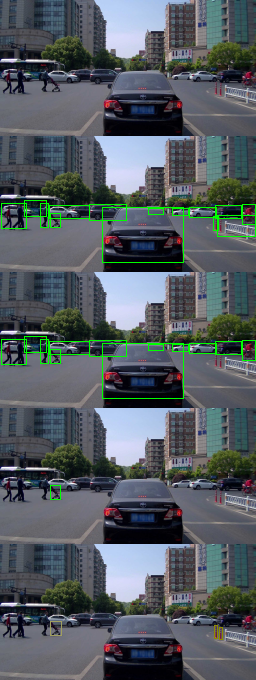}
        \includegraphics[width=0.1\linewidth]{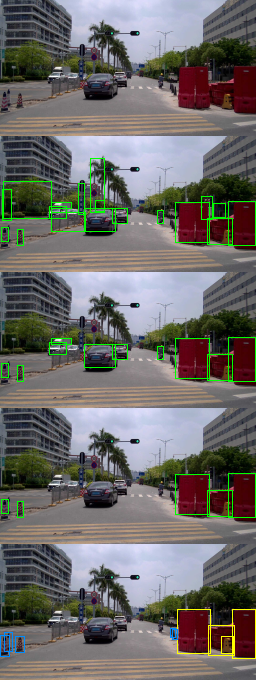}
        \includegraphics[width=0.1\linewidth]{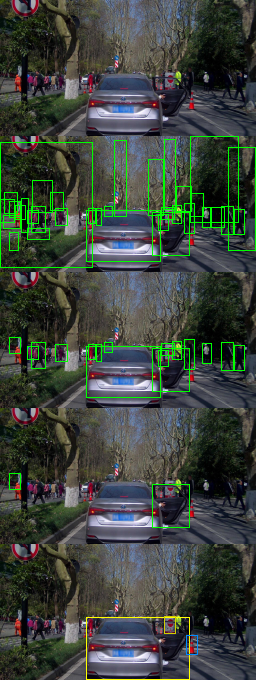}
    \end{minipage}
    \caption{Examples of COPG step-by-step proposals and final annotations. In each stack, the images from top to bottom correspond to the steps indicated in \cref{fig:proposal-pipeline}: \textbf{(b)} camera image, \textbf{(e)} initial proposal, \textbf{(f)} intermediate proposal, and \textbf{(g)} final proposal. The last row is the result after manual labeling. Best viewed with color and zoom in.}
    \label{fig:copg-examples}
\end{figure}

\end{document}